\documentclass[fleqn]{article} 
\usepackage{arxiv2019_conference,times}
\iclrfinalcopy

\usepackage{amsmath,amsfonts,bm}

\def\figref#1{figure~\ref{#1}}

\def\secref#1{section~\ref{#1}}

\def\eqref#1{equation~\ref{#1}}
\def\Eqref#1{Equation~\ref{#1}}

\def\1{\bm{1}}

\DeclareMathAlphabet{\mathsfit}{\encodingdefault}{\sfdefault}{m}{sl}
\SetMathAlphabet{\mathsfit}{bold}{\encodingdefault}{\sfdefault}{bx}{n}

\newcommand{\KL}{D_{\mathrm{KL}}}

\DeclareMathOperator*{\argmax}{arg\,max}

\usepackage{amsmath, amssymb}
\usepackage{hyperref,tikz}
\usepackage{url}

\usepackage{subfig}
\usepackage{graphicx,url}
\usepackage{floatrow}
\usepackage[export]{adjustbox}
\usepackage{wrapfig}
\usepackage{sidecap}
\usepackage{nccmath}
\usepackage{hyperref}
\usepackage{url}
\usepackage[utf8]{inputenc} 
\usepackage[T1]{fontenc}    
\usepackage{hyperref}       
\usepackage{url}            
\usepackage{booktabs}       

\usepackage{amsfonts}       
\usepackage{nicefrac}       
\usepackage{microtype}      
\usepackage{amsthm}

\usepackage{accents}

\newcommand{\pzo}{p_{0\rightarrow 1}}

\newcommand{\poz}{p_{1\rightarrow 0}}
\newcommand{\poo}{p_{1\rightarrow 1}}

\renewcommand{\argmax}{\arg\!\max}

\newcommand{\s}[1]{\tilde{#1}}
\newcommand{\hcm}{\hspace{0.5cm}}
\newcommand{\ocm}{\hspace{1cm}}

\newcommand{\ave}[2]{{\mathbb{E}_{#2}\sq{#1}}}
\newcommand{\david}[1]{}

\newcommand{\cb}[1]{\left\{ {#1} \right\}}
\newcommand{\br}[1]{\left( {#1} \right)}
\newcommand{\sq}[1]{\left[ {#1} \right]}

\renewcommand{\KL}[2]{\textrm{KL}\!\br{#1\vert\vert#2}}

\renewcommand{\div}[2]{\textrm{D}\!\br{#1\vert\vert#2}}
\newcommand{\fdiv}[2]{\textrm{D}_f\!\br{#1\vert\vert#2}}

\newcommand{\const}{{const.}}

\renewcommand{\figref}[1]{figure(\ref{#1})}
\renewcommand{\secref}[1]{section(\ref{#1})}

\renewcommand{\eqref}[1]{equation(\ref{#1})}
\renewcommand{\Eqref}[1]{Equation(\ref{#1})}

\newcommand{\beq}{\begin{equation}}
\newcommand{\eeq}{\end{equation}}

\newcommand{\ndist}[3]{{\cal{N}}\br{#1\thinspace\vline\thinspace #2,#3}}
\newcommand{\trans}{^{\textsf{T}}}

\newcommand{\ind}[1]{\mathbb{I}\br{#1}}
\newcommand{\qcm}{\hspace{0.2cm}}
\newcommand{\logreg}{\phi}

\title{Private Machine Learning via Randomised Response}

\author{David Barber\\\\
	Department of Computer Science\\
	University College London, UK
}

\begin{document}
	\maketitle

	\begin{abstract}
		We introduce a general learning framework for private machine learning based on randomised response. Our assumption is that all actors are potentially adversarial and as such we trust only to release a single noisy version of an individual's datapoint. 
		Our approach forms a consistent way to estimate the true underlying machine learning model and we demonstrate this in the case of logistic regression.
	\end{abstract}

\david{Whilst Differential Privacy is a different setting, it would still potentially be interesting to understand what guarantees the returned predictor has in terms of privacy. Is it possible to figure out whether any individual uploaded their noisy data to Facebook, based on the learned smile-detector that Facebook provides?}

\section{Private Machine Learning}

Our desire is to develop a strategy for machine learning driven by the requirement that private data should be shared as little as possible and that no-one can be trusted with an individual's data, neither a data collector/aggregator, nor the machine learner that tries to fit a model.


Randomised Response, see for example \cite{Warner65}, is relevant in this context in which a datapoint $x_n$ is replaced with a randomised `noisy' version $\tilde{x}_n$. A classical example is voting in an election in which an individual voter votes for one of two candidates $A$ or $B$ and is asked to lie (with probability $p$) about whom they voted for . This results in noisy data and estimating the fraction $f_A$ of voters that voted for candidate $A$ based on this noisy data 
\beq
\tilde{f}_A=\frac{1}{N}\sum_{n=1}^N \ind{\tilde{x}_n=A}
\label{eq:noisy:f}
\eeq
can give a potentially significantly incorrect estimate. As \cite{Warner65} showed, since we know the probabilistic mechanism that generated the noisy data, a better estimate of the fraction of voters voting for candidate $A$ is given by
\beq
f_A=\frac{\tilde{f}_A+p}{1-2p}
\label{eq:noisy:betterf}
\eeq

In a machine learning context, the kind of scenario we envisage is that users may have labelled face images as ``happy" or ``sad" on their mobile phones and the company MugTome wishes to train a ``happy/sad" face classifier; however, users do not wish to send the raw face images to MugTome and also wish to be able to plausibly deny which label they gave any training image. To preserve privacy, each user will send to MugTome only a single corrupted datapoint --- a single corrupted image and a single corrupted label.

It is straightforward to extend our approach to deal with users sending multiple corrupted datapoints. However, since MugTome will potentially then know which corrupted datapoints belong to each user, they will have more information to help reveal the underlying clean datapoint. Since we assume we cannot trust MugTome,  MugTome may attempt to recover the underlying true datapoint. For example, if a user sends three class labels $c_1,c_2,c_3$, $c_i\in\cb{0,1}$, then MugTome can have a good guess of the underlying true class label by simple taking the majority class $c=\ind{c_1+c_2+c_3 > 2}$. Indeed, in general, if $M$ corrupted datapoints are independently generated for a user, then MugTome's ability to reveal the true class (or attribute) increases dramatically. For example, if MugTome know the corruption mechanism $p(c_m|c_{true})$ the posterior of the class is given by
\beq
p(c_{true}|c_1,\ldots,c_M) \propto p(c_{true})\prod_{m=1}^M p(c_m|c_{true})
\eeq
where $p(c_{true})$ is the prior belief on the true class. This posterior distribution concentrates exponentially quickly (in $M$) around the true value $c_{true}$. Similarly, if a pollster asks each voter three times what they voted, then the questioner would have a very good idea of the true vote of each voter; to protect the voter's privacy, the voter would then have to trust that the pollster either does not pass on any information that states that the three votes came from the same person or that the pollster doesn't attempt themselves to figure out what the voter voted for. 

Similarly, in a medical setting in which a patient privately owns a datapoint, releasing $M$ synthetic versions (corruptions) of that datapoint can compromise privacy if which synthetic datapoints belong to each person is also known. To guarantee that privacy is retained would require patients to trust people with their data, namely that any data aggregation process will remove their patient ID. However, this is something out of the control of the patient and as such we do not consider generating multiple synthetic datapoints (see for example \cite{DBLP:journals/corr/abs-1708-07975}) a `safe' mechanism. 

For these reasons, we wish to make a process in which an individual only reveals a single corrupted datapoint; from that point onwards in the machine learning training process, no other trust in that process is required.  To motivate our general approach to private machine learning we discuss the voting example in more detail in \secref{sec:voting}. Connections to other forms of privacy preserving machine learning are discussed in \secref{sec:related}. The justification for our approach hinges on the properties of the Spread Divergence, which we review in the following section.


%
%
\section{Spread Divergence\label{sec:sd}}

Throughout we use the notation $p(X=x)$ for a random variable $X$ in state $x$. However, to reduce notational overhead, where unambiguous, we write simply $p(x)$.

A divergence  $\div{p}{q}$ (see, for example \citet{Dragomir2005}) is a measure of the difference between two distributions $p$ and $q$ with the property
\beq
\div{p}{q}\geq 0 \qcm\text{and} \qcm \div{p}{q} = 0 \qcm \Leftrightarrow \qcm p=q
\eeq
An important class is the $f$-divergence, defined as
\beq
\fdiv{p}{q} = \ave{f\br{\frac{p(x)}{q(x)}}}{q(x)}
\eeq
where $f(x)$ is a convex function with $f(1)=0$. A special case of an $f$-divergence is the well-known Kullback-Leibler divergence $\KL{p}{q} = \ave{\log{\frac{p(x)}{q(x)}}}{p(x)}$ which is widely used to train models using maximum likelihood. For the Spread Divergence, from $q(x)$ and $p(x)$ we define new distributions $\s{q}(\s{x})$ and $\s{p}(\s{x})$ that have the same support. Using the notation $\int_x$ to denote integration $\int\br{\cdot}dx$ for continuous $x$, and $\sum_{x\in {\cal X}}$ for discrete $x$ with domain ${\cal{X}}$, we define a random variable $\s{x}$ with the same domain as $x$ and distributions
\begin{align}
\s{p}(\s{x}) = \int_x p(\s{x}{\mid}x)p(x), \hcm \s{q}(\s{x}) = \int_x p(\s{x}{\mid}x)q(x)
\end{align}
where $p(\s{x}{\mid}x)$ `spreads' the mass of $p$ and $q$ such that $\s{p}(\s{x})$ and $\s{q}(\s{x})$ have the same support. For example, if we use a Gaussian $p(\s{x}{\mid}x)=\ndist{\s{x}}{x}{\sigma^2}$, then $\s{p}$ and $\s{q}$ both have support $\mathbb{R}$. The spread divergence has a requirement on the noise $p(\s{x}|x)$, namely that $\div{\s{p}}{\s{q}} = 0 \Leftrightarrow p=q$; that is, if the divergence of the spreaded distributions is zero, then the original non-spreaded distribution will match.  As shown in \cite{zhang2018spread} this is guaranteed for certain `spread noise' distributions. In particular, for continuous $x$ and $\s{x}$ of the same dimension and injective function $f$,  a sufficient condition for a valid spread noise $p(\s{x}|x)=K(\s{x}-f(x))$ is that the kernel $K(x)$ has strictly positive Fourier Transform. For discrete variables, a sufficient condition is that $p(\s{x}=i|x=j)=P_{ij}$ is that $P_{ij}>0$ and the matrix $P$ is square and invertible. 
%
%




Spread divergences have a natural connection to privacy preservation and Randomised Response \citep{Warner65}. The spread divergence suggests a general strategy to perform private machine learning. We first express the machine learning problem as $\min_\theta\fdiv{{p}(X)}{{p}_\theta(X)}$ for a specified model $p_\theta(X)$. Then, given only noisy data $\s{X}$, we fit the model by $\min_\theta\fdiv{\s{p}(\s{X})}{\s{p}_\theta(\s{X})}$.  To explain in more detail how this works, we first describe randomised response in a classical voting context and then justify how to generalise this to principled training of machine learning models based on corrupted data.

\section{A classical voting example\label{sec:privacy:voting}\label{sec:voting}}

There are two candidates in an election, candidate ``one" and candidate ``zero" and Alice would like to know the fraction of voters that voted for candidate ``one".  We write the dataset of voting as a collection of binary values $\cb{x_1,\ldots, x_N}$, $x_n\in\cb{0,1}$.  

\subsection{Learning $\theta$ using clean data}

If we assume that Alice has full knowledge of which candidate each voter voted for, then  clearly Alice may simply count the fraction of people that voted for ``one'' and set
\beq
\theta= \frac{1}{N}\sum_{n=1}^N x_n
\label{eq:frac:one}
\eeq
It will be useful to first consider how to arrive at the same result from a modelling perspective. We can consider an independent Bernoulli model
\beq
p_\theta(X_1=x_1,\ldots,X_N=x_N) = \prod_{n=1}^N p_\theta(X_n=x_n)
\eeq
where
\beq
p_\theta(X=1) = \theta
\eeq
so that 
\beq
p_\theta(X=x) = \theta^x\br{1-\theta}^{1-x}
\eeq
We also construct an empirical data distribution that places mass only on the observed joint state, namely 
\beq
\hat{p}(X_1,\ldots,X_N)=\prod_{n=1}^N \delta(X_n,x_n) 
\eeq
where $\delta(x,x')$ is the Kronecker delta function. Then
\beq
\frac{1}{N}\KL{\hat{p}(X_1,\ldots,X_N)}{p_\theta(X_1,\ldots,X_N)}=L_N(\theta) + \const
\eeq
where
\begin{align}
L_N(\theta) &= \frac{1}{N}\sum_{n=1}^N \log p_\theta(X_n=x_n)\\
&=\frac{1}{N}\sum_{n=1}^N \br{x_n\log\theta+(1-x_n)\log\br{1-\theta}}
\end{align}
and minimising $\KL{\hat{p}}{p_\theta}$ (or maximising $L_N(\theta)$) with respect to $\theta$ recovers the fraction of votes that are 1, \eqref{eq:frac:one}. This shows how we can frame estimating the quantity $\theta$ from uncorrupted private data as a divergence minimisation problem.

\subsection{Learning $\theta$ using corrupted data}

Returning to the privacy setting, Bob would also like to know the fraction of votes that are 1. However, Alice does not want to send to Bob the raw data $x_1,\ldots,x_N$ since the votes of any individual should not be explicitly revealed. To preserve privacy, Alice sends noisy data ${\s{x}_1,\ldots,\s{x}_N}$ to Bob. In this case we draw a single joint sample $\s{x}_1,\ldots,\s{x}_N$ from the distribution
\beq
p(\s{X}_1,\ldots,\s{X}_N|X_1,\ldots,X_N)=\prod_{n=1}^N p(\s{X}_n|X_n)
\eeq
where the `spread noise' model is $p(\s{X}_n=i|X_n=j)=P_{ij}$. Hence, if $x_n=0$ Alice draws a sample $\s{x}_n=0$ with probability $P_{00}$ and $\s{x}_n=1$ with probability $P_{10}$. 

Given a sampled noisy dataset $\s{x}_1,\ldots,\s{x}_N$ we form an empirical spreaded data distribution 
\beq
\hat{p}(\s{X}_1,\ldots,\s{X}_N)=\prod_{n=1}^N \delta\br{\s{X}_n,\s{x}_n}
\eeq
Similarly, the corrupted joint model is given by
\beq
\s{p}_\theta(\s{X}_1,\ldots,\s{X}_N)=\prod_{n=1}^N \s{p}_\theta(\s{X}_n)
\eeq
where 
\beq
\s{p}_\theta(\s{X}=j) = \sum_{j\in\cb{0,1}} p(\s{X}=j|X=j)p_\theta(X=j)
\eeq
On receiving the noisy dataset $\s{x}_1,\ldots,\s{x}_N$, Bob can try to estimate $\theta$ by minimising 
\beq
\frac{1}{N} \KL{\hat{p}(\s{X}_1,\ldots,\s{X}_N)}{p_\theta(\s{X}_1,\ldots,\s{X}_N)}=-\frac{1}{N}\sum_{n=1}^N \log \s{p}_\theta(\s{x}_n) + \const
\label{eq:obj}
\eeq
with respect to $\theta$. Equivalently, he may maximise the scaled spread log likelihood
\beq
\s{L}_N(\theta)= \frac{1}{N}\sum_{n=1}^N \log \s{p}_\theta(\s{x}_n)
\label{eq:slik}
\eeq
For this simple model, Bob can easily explicitly calculate 
\beq
\s{p}_\theta(\s{x}=1)=p(\s{x}=1|x=1)p_\theta(x=1) + p(\s{x}=1|x=0)p_\theta(x=0)=P_{11}\theta+P_{10}\br{1-\theta}
\eeq
Similarly, $\s{p}_\theta(\s{x}=0)=P_{01}\theta+P_{00}\br{1-\theta}$.  In this case, \eqref{eq:slik} becomes
\beq
\s{f}_0\log\br{P_{00}\br{1-\theta}+P_{01}\theta}+\s{f}_1\log\br{P_{10}\br{1-\theta}+P_{11}\theta}
\eeq
where
\beq
\s{f}_1 = \frac{1}{N}\sum_{n=1}^N \s{x}_n
\eeq
Using $\s{f}_0+\s{f}_1=1$, $P_{00}+P_{10}=1$,  $P_{01}+P_{11}=1$, the maximum of the spread log likelihood is at
\beq
\theta = \frac{\s{f}_1+P_{10}}{1-P_{10}-P_{01}}
\eeq
which forms Bob's estimate of the underlying fraction of voters that voted for candidate ``one''. 


For example, if there were no noise $P_{10}=P_{01}=0$, Bob would estimate $\theta = \s{f}_1$, simply recovering the fraction of votes that are 1 in the original data. In the limit of a large number of votes $N\rightarrow\infty$ and true probability $\theta_0$ of a voter voting for candidate ``one'', then $\tilde{f}_0$ tends to $P_{00}(1-\theta_0)+P_{01}\theta_0$ and Bob's estimate recovers the true underlying voting probability $\theta=\theta_0$. Hence, even though Bob only receives a corrupted set of votes, in the limit of a large number of votes, he can nevertheless estimate the true fraction of people that voted for candidate ``one''.

\section{Private Machine Learning using Randomised Response\label{sec:pml}}

The above example suggests a general strategy to perform private machine learning: 

\begin{enumerate}
	\item{Phrase problem as likelihood maximisation:}
We first assume that a machine learning task for private data $x_1,\ldots, x_N$ can be expressed as learning a data model $p_\theta(X)$ by optimising an objective
\beq
L_N(\theta)= \frac{1}{N}\sum_{n=1}^N \log p_\theta(X_n=x_n)
\eeq

\item{Form a corrupted dataset:} Draw a single joint sample $\s{x}_1,\ldots,\s{x}_N$ from the distribution
\beq
p(\s{X}_1,\ldots,\s{X}_N|X_1,\ldots,X_N)=\prod_{n=1}^N p(\s{X}_n|X_n)
\eeq
where $p(\s{X}|X)$ is a defined spread noise distribution and known by both the owner of the private data and the receiver of the corrupted data. To do this, we go through each element of the dataset $x_n$ and replace it with a corruption $\s{x}_n$ sampled from $p(\s{X}_n=\s{x}_n|X_n=x_n)$. 

\item{Send data to learner:} We then send to the learner the corrupted dataset $\s{x}_1,\ldots,\s{x}_N$, the model to be learned $p_\theta(X)$ and the corruption probability $p(\s{X}|X)$.

\item{Estimate $\theta$ from corrupted data:} Having received the corrupted data $\s{x}_1,\ldots,\s{x}_N$, the learner fits $\theta$ by maximising the objective
\beq
\s{L}_N(\theta)=\frac{1}{N}\sum_{n=1}^N \log \s{p}_\theta(\s{x}_n) 
\label{eq:ob}
\eeq
where
\beq
\s{p}_\theta(\s{x})= \int_x p(\s{x}|x)p_\theta(x)	
\eeq

\end{enumerate}

\subsection{Justification}
%
%
%
%
%

%

If we assume that each element $x_n$ of the training data $x_1,\ldots,x_N$ is identically and independently sampled from a model $p_{\theta_0}(X_n=x_n)$, then each corrupted observation $\s{x}_n$ is a sample from the same distribution given by
\beq
\s{p}_{\theta_0}(\s{X}_N=y_n) = \int_x p(\s{X}_n=y_n|X=x)p_{\theta_0}(X=x)
\eeq
By the law of large numbers the objective \eqref{eq:ob} approaches its average over the data generating mechanism \david{what about the finite variance requirement? I guess if the spread noise is valid then the probability $ \s{p}_\theta(Y=y_n)$ cannot be zero so that the log probability is bounded and the variance is also bounded}
\beq
\lim_{N\rightarrow\infty}L_N(\theta)\xrightarrow{a.s.} \int_{\s{x}} \s{p}_{\theta_0}(\s{X}=\s{x}) \log \s{p}_\theta(\s{X}=\s{x})
\eeq
and maximising the spread likelihood objective $\s{L}_N(\theta)$ becomes equivalent to minimising
\beq
\KL{\s{p}_{\theta_0}(\s{X})}{\s{p}_\theta(\s{X})}
\eeq
Provided that the spread noise is valid (see \secref{sec:sd}), then
\beq
\KL{\s{p}_{\theta_0}(\s{X})}{\s{p}_\theta(\s{X})} = 0 \Rightarrow \theta=\theta_0
\eeq
for an identifiable model $p_\theta$. Thus 
\beq
\theta_{est} = \argmax_{\theta} \s{L}_N(\theta)
\eeq
is a consistent estimator. 

This means that (in the large data limit and assuming the training data is generated from the model), even though we only train on corrupted data, we are optimising an objective $\s{L}_N(\theta)$ which has a global minimum close to that of the objective on uncorrupted data $L_N(\theta)$.  Indeed, the estimator is consistent in the sense that as the amount of training data increases, we will recover the true clean data generating mechanism.  Hence, provided that the corruption process is based on spread noise, then we can still learn the model parameters $\theta$ even by training on only corrupted data.  In our motivating voting scenario in \secref{sec:voting}, we saw explicitly that the estimate $\theta$ of the true underlying voting fraction is consistent and indeed, this is a general property of our approach. 

\subsection{Training on noise only}

A common approach in private machine learning is to form synthetic (noisy, corrupted) data and then simply train the standard model on this noisy data --- see for example \cite{li2019deepobfuscator}. In our notation, this would be equivalent to maximising the likelihood
\beq
L'_N(\theta)\equiv \frac{1}{N}\sum_{n=1}^N \log p_\theta(X=\s{x}_n)
\eeq
As above, assuming that the training data is generated from an underlying model $p_{\theta_0}(X_n=x_n)$, by the law of large numbers, 
\beq
\lim_{N\rightarrow\infty}L'_N(\theta)\xrightarrow{a.s.} \int_{\s{x}} \s{p}_{\theta_0}(\s{X}=\s{x}) \log p_\theta({X}=\s{x})
\label{eq:log:lik:noise}
\eeq
In general, the optimum of this objective does not occur when $\theta=\theta_0$ and therefore training on noisy data alone does not form a consistent estimator of the true underlying model. 

We discuss learning with noisy labels more extensively in the context of logistic regression in \secref{sec:train:on:noise:only} in which we show that provided the label flip noise is not too high $\pzo+\poz<1$, and for zero mean isotropically Gaussian distributed inputs, maximum likelihood training with corrupted class labels does form a consistent estimator. Hence, whilst one cannot guarantee that maximum likelihood training of logistic regression on noisy data will result in a consistent estimator, there are special situations in which this may work.


%
%
%
%

\begin{figure}[t]
	\centering
	\subfloat[$p_f=0$]{\includegraphics[width=0.17\linewidth]{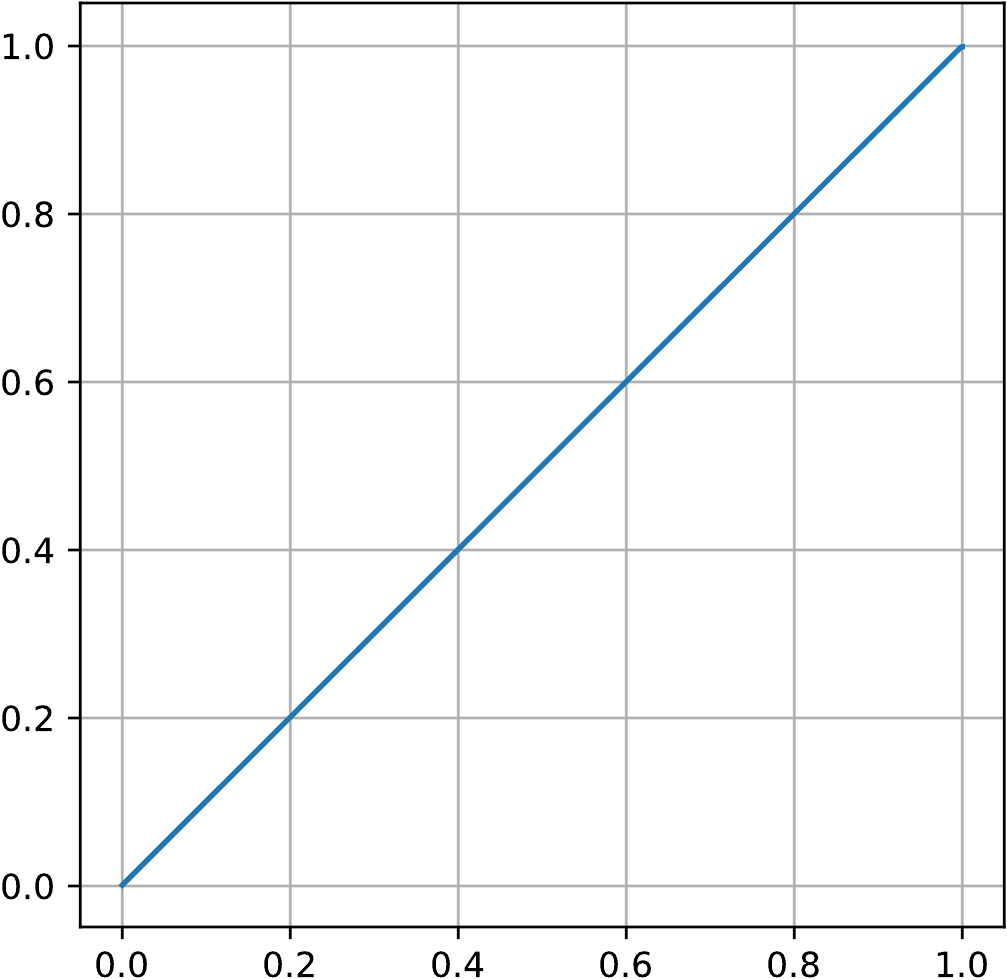}}\hspace{2mm}
	\subfloat[$p_f=0.001$]{\includegraphics[width=0.17\linewidth]{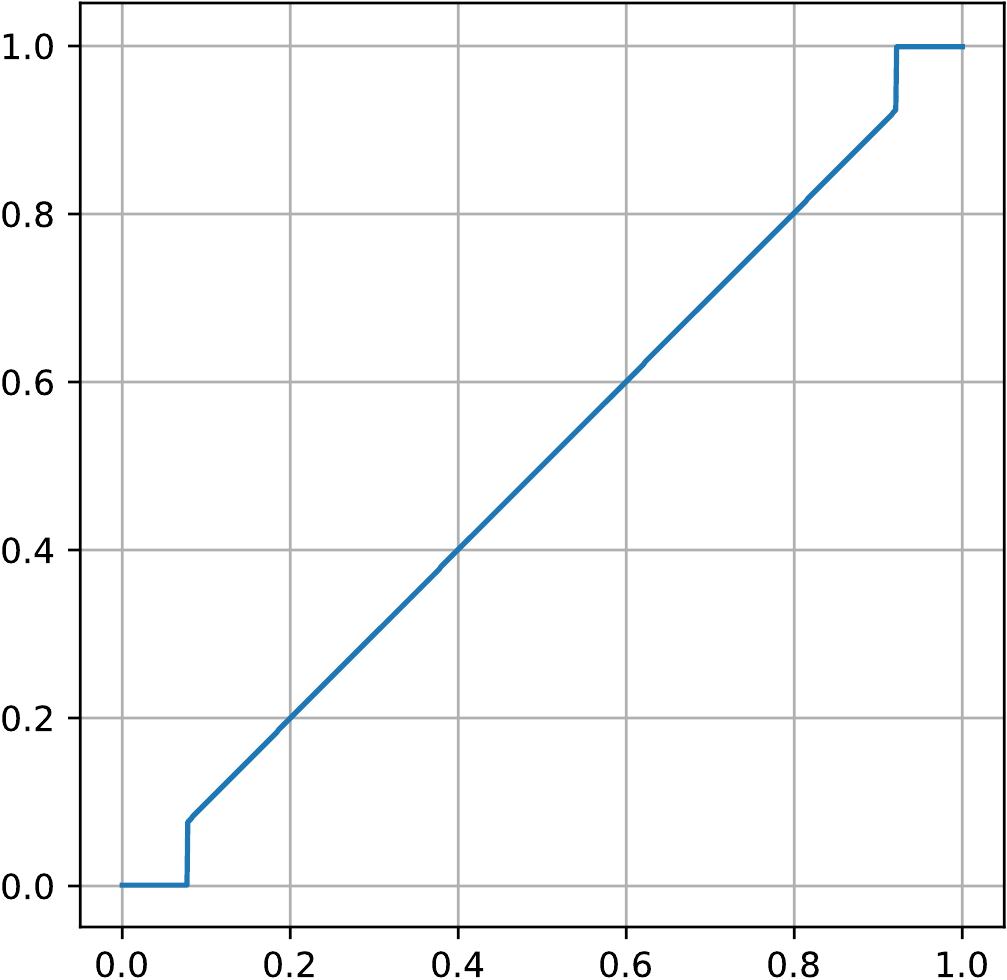}}\hspace{2mm}
	\subfloat[$p_f=0.002$]{\includegraphics[width=0.17\linewidth]{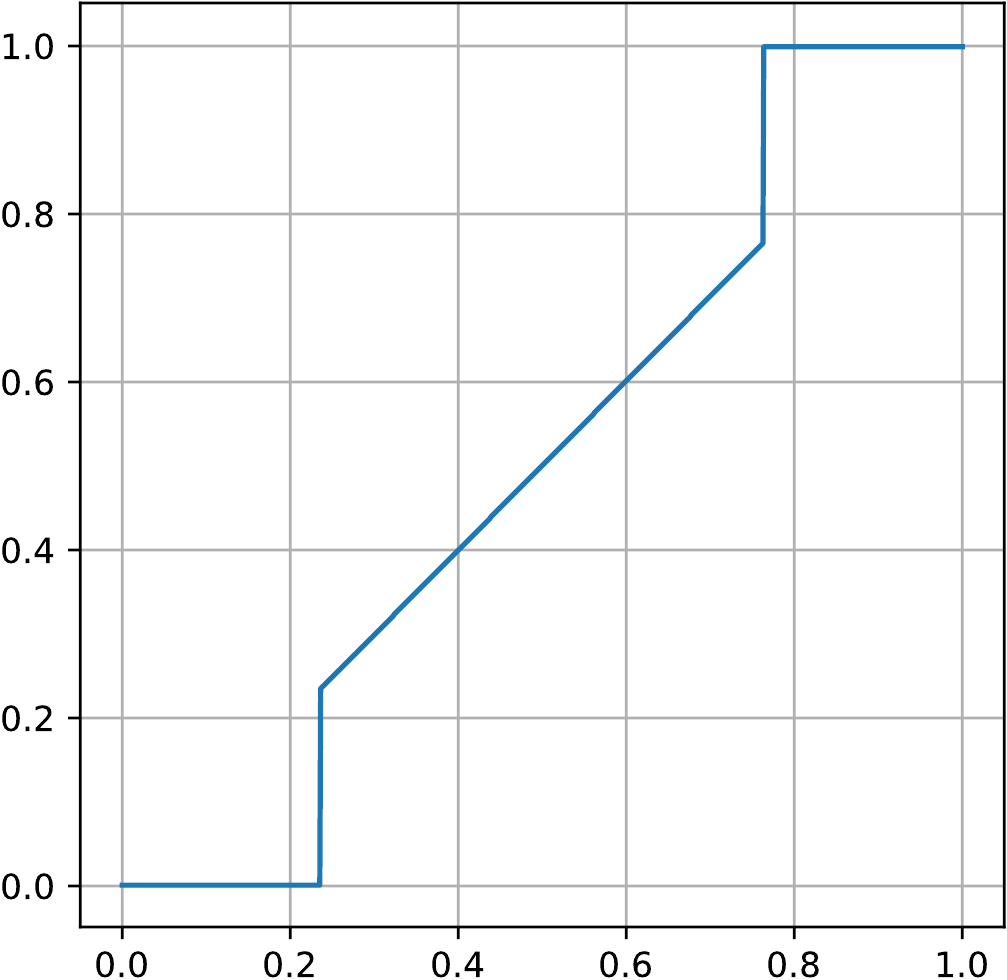}}\hspace{2mm}
	\subfloat[$p_f=0.003$]{\includegraphics[width=0.17\linewidth]{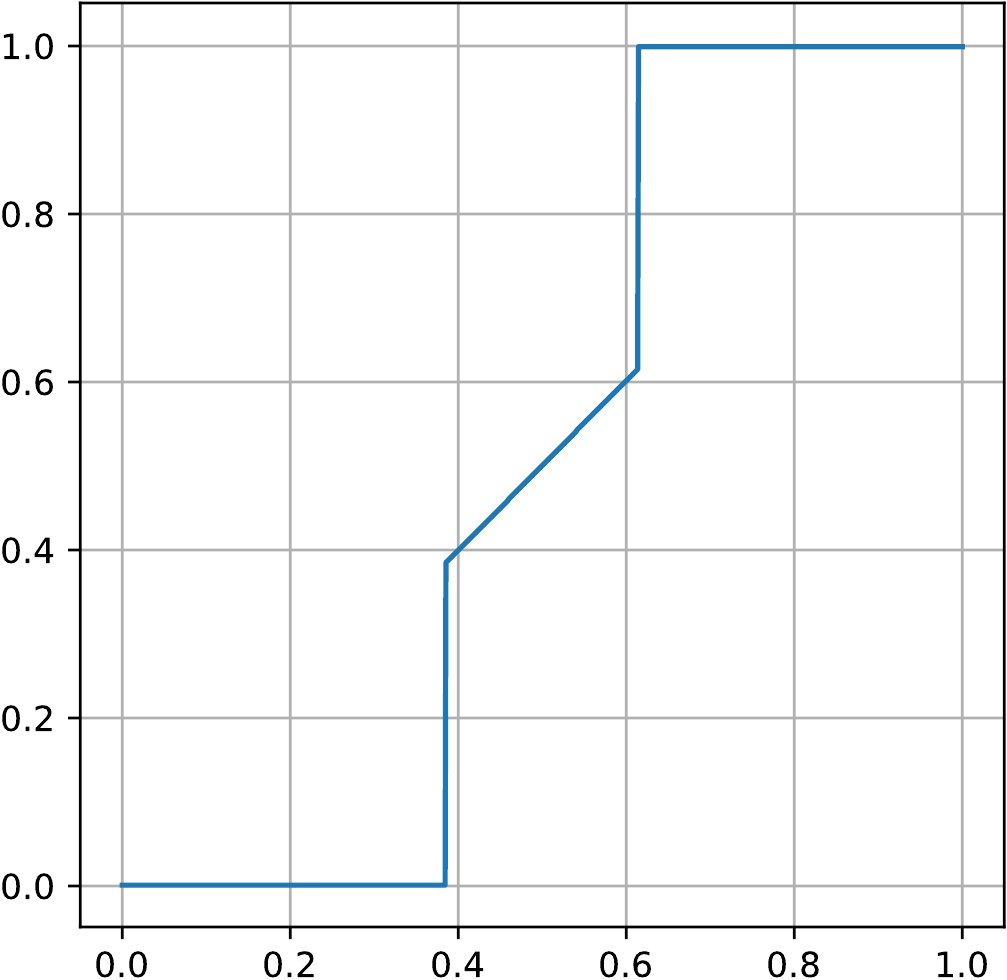}}\hspace{2mm}
	\subfloat[$p_f=0.004$]{\includegraphics[width=0.17\linewidth]{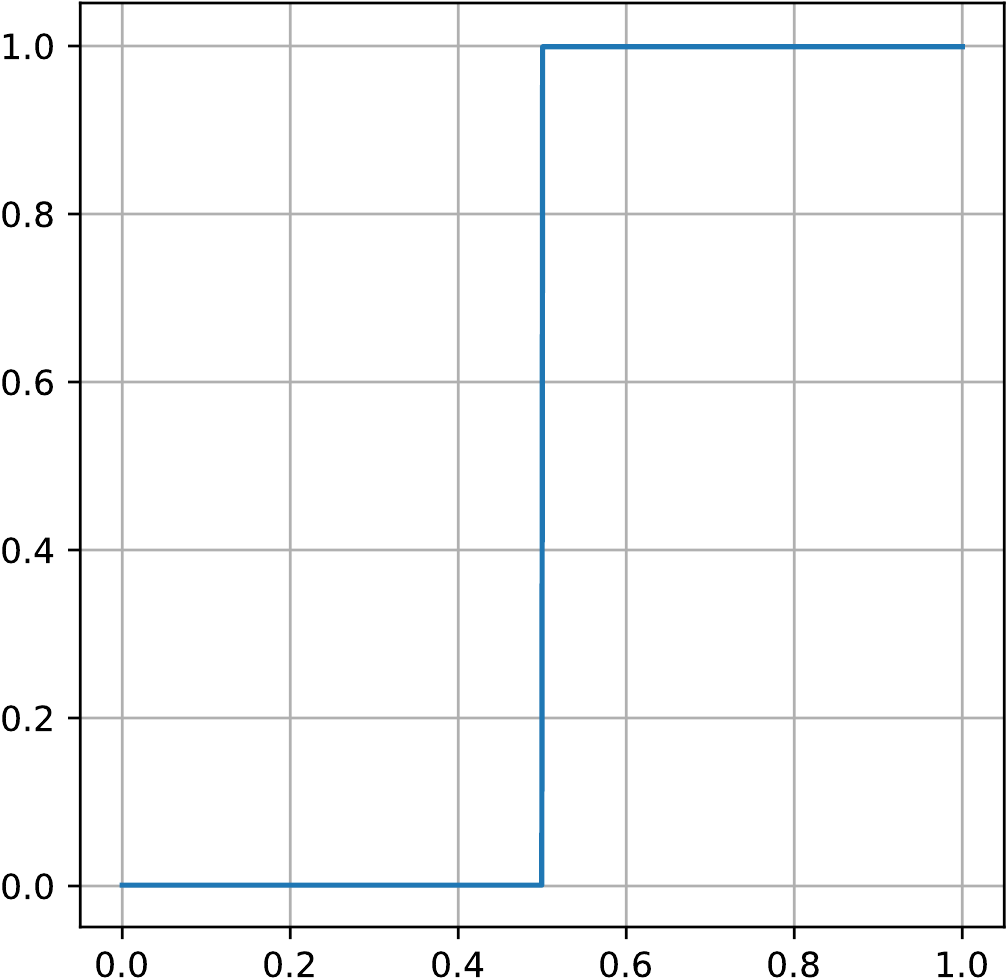}}
	\caption{Training based on the reconstruction approach, \secref{sec:recon} for the model with binary variable $p(x=1)=\theta$, $p(x=0)=1-\theta$. In each case we plot along the $x$-axis the true $\theta_0$ from 0 to 1 and on the $y$-axis the value of $\theta$ that maximises $J_\infty(\theta)$. In each plot we use a different flip probability. For a consistent estimator we would require that each plot is a straight $x=y$ line, which only occurs in the case of no noise, $p_f=0$. \label{fig:opt}}
\end{figure}

\subsection{Reconstruction Approach\label{sec:recon}}

A seemingly natural alternative to our method is to attempt to reconstruct the clean datapoint from the noisy datapoint and use that within a standard learning framework. This approach would give an objective
\beq
J_N(\theta) = \frac{1}{N}\sum_{n=1}^N \int_{x_n} p(x_n|\s{x}_n)\log p_\theta(x_n) 
\eeq
Here we need to define a posterior distribution $p(x_n|\s{x}_n)$ to reconstruct the clean datapoint. Since the learner only has knowledge of the prior $p_\theta(x)$ it is natural to set
\beq
p(x_n|\s{x}_n) = p_\theta(x_n|\s{x}_n) \equiv \frac{p(\s{x}_n|x_n)p_{\theta}(x_n)}{\int_{x_n} p(\s{x}_n|x_n)p_{\theta}(x_n)}
\eeq
By the law of large numbers $J_N$ converges to its expectation with respect to the true data generating mechanism $p_{\theta_0}(\s{x})= \int p(\s{x}|x)p_{\theta_0}(x)$, so that 
\begin{align}
\lim_{N\rightarrow\infty}J_N(\theta)\xrightarrow{a.s.} \int_{x,\s{x}} p_{\theta_0}(\s{x}) p_\theta(x|\s{x})\log p_\theta(x)\equiv J_\infty(\theta)
\end{align}
In general, the optimum of $J_\infty(\theta)$ is not at $\theta=\theta_0$. To demonstrate this, we plot in \figref{fig:opt} the optimal $\theta$ for a simple Bernoulli model for which we can calculate $J_\infty(\theta)$ exactly. As we see, for all but zero flip noise, $p_f=0$, the estimator does not correctly identify the underlying probabilty generating mechanism. For this reason, we do not pursue this approach further. 

\subsection{Other Divergences}

An extension of the above is to learn $\theta$ by minimising other $f$-divergences 
\beq
\fdiv{\s{p}_\theta(Y)}{\hat{p}(Y|y)}= \ave{f\br{\frac{\s{p}_\theta(Y)}{\hat{p}(Y|y)}}}{\hat{p}(Y|y)}
\eeq
However, this generalisation to any $f$-divergence is harder to justify since the expectation of this objective (by averaging over the noise realisations)
\beq
\int \s{p}(y)\fdiv{\s{p}_\theta(Y)}{\hat{p}(Y|y)}
\eeq
will not in general give a divergence between spreaded distributions. This means that in the limit of a large number of datapoints, it is not guaranteed to recover the true data generating process, except for special choices of the $f$-divergence, such as the KL divergence. We leave a discussion of this for future work. \david{Are there other divergences or divergence classes for which this might work?}

\section{Private Logistic Regression\label{sec:private:log:reg}}

As an application of the above framework to a standard machine learning model, we now discuss how to form a private version of logistic regression.

Returning to our motivating example, users may have labelled face images as ``happy" or ``sad" on their mobile phones and the company MugTome wishes to train a ``happy/sad" face classifier; however, users do not wish to send the raw face images to MugTome and also wish to be able to plausibly deny which label they gave any training image. 

In this case we have a set of training data $x_1,\ldots,x_N$, $x_n\in\mathbb{R}^D$ and corresponding binary class labels $c_1,\ldots,c_N$, $c_n\in\cb{0,1}$. We wish to fit a logistic regression model

\beq
p_\theta(c|x) = \logreg( (2c-1)\theta_c\trans x)
\eeq 
where $\logreg(x)=1/(1+e^{-x})$ is the logistic function. We follow the general approach outlined in \secref{sec:pml}.

\begin{enumerate}
	\item{The model:} For observation $(x,c)$ and parameter $\theta$
\beq
p_{\theta}(c,x)=p_{\theta_c}(c|x)p_{\theta_x}(x)
\eeq
where $p_{\theta_c}(c|x)$ is the standard logistic regression model above and $p_{\theta_x}(x)$ is a model of the input $x$. The training objective is
\begin{align}
L_N(\theta)& =\frac{1}{N}\sum_{n=1}^N \log p_{\theta}(c_n,x_n)\\
&=\frac{1}{N}\sum_{n=1}^N \log p_{\theta_c}(c_n|x_n) + \frac{1}{N}\sum_{n=1}^N \log p_{\theta_x}(x_n)
\label{eq:logistic:lik}
\end{align}
We note that this is a separable objective for $L_N(\theta)=L^c_N(\theta_c)+L^x_N(\theta_x)$, in which the logistic regression parameters $\theta_c$ are conditionally independent (conditioned on the training data) of the input parameters $\theta_x$.

\item{Form the corrupted dataset:} We wish to send noisy data $\s{x}_1,\ldots,\s{x}_N$, $\s{c}_1,\ldots,\s{c}_N$ to the learner. To do so we need to define a corruption model $p(\s{c},\s{x}|c,x)$. For simplicity, we consider a corruption model of the form
\beq
p(\s{c},\s{x}|c,x)=p(\s{c}|c)p(\s{x}|x)
\eeq
The corruption processes of $p(\s{c}|c)$ and $p(\s{x}|x)$ are problem specific; see the experiments \secref{sec:exp} for some examples.

\item{Send to learner corrupted data and model:} The corrupted labels and inputs are sent to the learner $(\tilde{c}_1,\tilde{x}_1),\ldots,(\tilde{c}_N,\tilde{x}_N)$ along with the model $p_{\theta_c}(c|x)$, $p_{\theta_x}(x)$ and corruption process $p(\tilde{c}|c)$, $p(\tilde{x}|x)$.

\item{Learn the model parameters $\theta$:} The spread log likelihood is 
\begin{align}
\s{L}(\theta) &=\frac{1}{N}\sum_{n=1}^N \log \tilde{p}_{\theta}(\tilde{c}_n,\tilde{x}_n)\\
&=\frac{1}{N}\sum_{n=1}^N \log \int_{x_n,c_n} p(\tilde{c}_n|c_n) p(\tilde{x}_n|x_n) p_{\theta_c}(c_n|x_n)p_{\theta_x}(x_n)
\end{align}
Unfortunately, in all but special cases, the integral (for continuous $x$) or sum (for discrete $x$) required to evaluate $\s{L}$ is not tractable and numerical approximation is required. For this stage, there are many options available and we present below the approach taken in the experiments.

Interestingly, we note that, unlike training on clean data, the objective $\s{L}(\theta)$ is not separable into a function of $\theta_c$ plus a function of $\theta_x$, meaning that learning the class prediction parameter $\theta_c$ is coupled with learning the input distribution parameter $\theta_x$. 

\end{enumerate}

\subsection{Implementation}

In general, the spread noise defines a distribution on a pair of spread variables $p(\s{c},\s{x}|c,x)$ and the full joint distribution, including the original model is
\beq
p(\s{c},\s{x},c,x) = p(\s{c},\s{x}|c,x)p_{\theta_c}(c|x)p_{\theta_x}(x)
\eeq
For continuous $x$, the spread likelihood is then obtained from
\beq
p(\s{c},\s{x}) = \sum_c \int_x p(\s{c},\s{x},c,x)
\eeq
In general, this sum/integral over $x$ is intractable due to the high-dimensionality of $x$. We use a standard approach to lower bound the log likelihood (for a single datapoint) by
\begin{multline}
\log p(\s{c},\s{x}) \geq -\ave{\log q(c,x|\s{c},\s{x})}{q(c,x|\s{c},\s{x})}+\ave{\log p(\s{c},\s{x}|c,x)p_{\theta_c}(c|x)p_{\theta_x}(x)}{q(c,x|\s{c},\s{x})}
\end{multline}
where $q$ is a distribution chosen to make the bound tight, see for example \cite{Barber:2012:BRM:2207809}. This allows us to use an EM-style procedure in which we iterate between the two steps : (M-step) fix $q$ and optimise $\theta$ and (E-step) fix $\theta$ and update $q$. 

\begin{enumerate}
	\item{Iteration $k$ M-step:} Update $\theta$ to increase the ``energy''
	\beq
	\theta^{k+1} = \argmax_{\theta} E(\theta; q_k)
	\eeq
	where (for multiple datapoints)
\beq
E(\theta;q) \equiv \sum_{n=1}^N \ave{\log p_{\theta_c}(c_n|x_n)}{q(c_n,x_n|\s{c}_n,\s{x}_n)}
+\sum_{n=1}^N \ave{\log p_{\theta_x}(x_n)}{q(x_n|\s{c}_n,\s{x}_n)}
\label{eq:logistc:energy}
\eeq
An advantage of this approach is that $E(\theta;q)$ is separable and we can update the class prediction parameter $\theta_c$ independently of the input distribution parameter $\theta_x$. 

In practice we will typically only do a partial optimisation (gradient ascent step) over $\theta$ to guarantee an increase in the energy.

\item{Iteration $k$ E-step:} 
The bound is tightest when $q$ is set to the posterior (see for example \citet{Barber:2012:BRM:2207809}),  
\beq
q_{k+1}(c,x|\s{c},\s{x})=p(c,x|\s{c},\s{x}) = \frac{p(\s{c}|c)p(\s{x}|x)p(c|x)p(x)}{Z(\s{c},\s{x})}
\eeq
where $p(c|x) = p_{\theta^k_c}(c|x)$, $p(x) = p_{\theta^k_x}(c|x)$ and the normaliser is given by
\beq
Z(\s{c},\s{x}) \equiv \sum_c \int  p(\s{c}|c)p(\s{x}|x)p(c|x)p(x)dx
\eeq
\end{enumerate}
To implement the M-step, \Eqref{eq:logistc:energy} requires expectations of the form
\beq
\sum_c\int  p(c,x|\s{c},\s{x}) f(x,c) dx
\eeq
for some function $f(x,c)$. Assuming that the posterior will be reasonably peaked around the noisy data we use sampling with an importance distribution
\beq
\rho(c,x|\tilde{c},\tilde{x})=\rho(c|\tilde{c})\rho(x|\tilde{x})
\eeq
The expectation is then motivated by
\beq
\sum_c\int_x p(c,x|\s{c},\s{x}) f(x,c)= \sum_c\int_x  \rho(c|\s{c})\rho(x|\s{x}) \frac{p(\s{c}|c)p(\s{x}|x)p(c|x)p(x)}{\rho(c|\s{c})\rho(x|\s{x})Z(\s{c},\s{x})}f(x,c)
\eeq
Choosing
\beq
\rho(c|\s{c}) = \frac{p(\s{c}|c)}{Z_\rho(\s{c})}, \hcm \rho(x|\s{x}) = \frac{p(\s{x}|x)p(x)}{Z_\rho(\s{x})}
\eeq
for normalising functions $Z_\rho(\s{c})$, $Z_\rho(\s{x})$ we then run a standard importance sampling approximation (see \secref{app:logreg}). For a given noisy datapoint $(\s{c}_n,\s{x}_n)$ we generate a set of $S$ samples $c_n^1,\ldots, c_n^S$ from $\rho(c|\s{c}_n)$ and samples $x_n^1,\ldots, x_n^S$ from $\rho(x|\s{x}_n)$ and compute the importance weights 
\beq
w(s|n) = \frac{\logreg\br{ (2c_n^s-1)\theta_c\trans{}x_n^s}}{\sum_s \logreg\br{ (2c_n^s-1)\theta_c\trans{}x_n^s}}
\eeq
The energy \eqref{eq:logistc:energy} separates into two independent terms (see \secref{app:logreg})
\beq
E(\theta_c;q)\approx \sum_{n=1}^N\sum_{s=1}^S   w(s|n)\log\logreg\br{ (2c_n^s-1)\theta_c\trans{}x_n^s}
\label{eq:E:c}
\eeq
and
\beq
E(\theta_x;q)\approx \sum_{n=1}^N\sum_{s=1}^S  w(s|n)\log p_{\theta_x}(x_n^s)
\label{eq:E:x}
\eeq
\Eqref{eq:E:c} is a weighted version of the standard logistic regression log likelihood, $L^c(\theta_c)$ in \eqref{eq:logistic:lik}; similarly \eqref{eq:E:x} is a weighted version of $L^x(\theta_x)$. The advantage therefore is that, given the importance samples, the learning procedure for $\theta$ requires only a minor modification of the standard maximum likelihood training procedure on clean data.

The full procedure is that we randomly initialise $\theta$ and then, for each datapoint $n$, draw samples and accumulate the gradient across samples and datapoints. After doing a gradient ascent step in $\theta$, we update the importance distributions and repeat until convergence. 

The Importance Sampling approximation is a convenient approach, motivated by the assumption that corrupted datapoints will be close to their uncorrupted counterparts. Whilst we used a bound as part of the approximation, this is not equivalent to using a parametric $q$ distribution; by sampling we form a consistent estimator of the tightest possible lower bound. In other words, we are simply using Importance Sampling to estimate the expectations required within a standard Expectation Maximisation algorithm, see for example \cite{Barber:2012:BRM:2207809}. We also tried learning a parametric $q$, similar to standard variational approaches to log likelihood maximisation, but didn't find any improvement on the Importance Sampling approach.

\subsection{Learning the prior}

If we have access to clean data, the optimal input model $p_{\theta_x}(x)$ can be learned from maximising the likelihood $L^x(\theta_x)$. However, our general assumption is that we will never have access to clean data.  
There may be situations in which the learner has a good model of  $p_{\theta_x}(x)$, without compromising privacy (for example a publicly available dataset for a similar prediction problem might be available) in which case it makes sense to set the prior to this known model.

In the absence of a suitable prior  we can attempt to learn $p_{\theta_x}(x)$ from the corrupted data by maximising $\s{L}(\theta)$.
For simplicity we assume a factorised model and for a $D$-dimensional input vector $x=(x[1],\dots,x[D])$ write
\beq
p_{\theta_x}(x) = \prod_{d=1}^D p(x[d]|d)
\eeq
for a collection of learnable univariate distributions $p(x[d]|d)$, $d=1,\ldots,D$. Under this assumption, and using the Importance Sampling approach in \eqref{eq:E:x}, this means that $p(x[d]|d)$ can be learned by maximising
\beq
E_x = \sum_{n=1}^N\sum_{s=1}^S w(s|n)\sum_{d=1}^D \log p(x^n_s[d]|d)
\eeq
Since this is a separable objective, we can learn each $p(x^n_s[d]|d)$ independently. 

For simplicity, we assume a discrete distribution for $x[d]$ that contains $K$ states (or bins). Then
\beq
E_x[d] = \sum_{k=1}^K \sum_{n=1}^N\sum_{s=1}^S  w(s|n) \ind{x^n_s[d] \in k}\log p(k|d)
\eeq
where $ \ind{x^n_s[d] \in k}$ is 1 if the sample $x^n_s[d]$ is in the $k^{th}$ state and 0 otherwise. Optimising with respect to $p(k|d)$ we obtain
\beq
p(k|d) = \frac{{\sum_{n=1}^N\sum_{s=1}^S  w(s|n) \ind{x^n_s[d] \in k}}}{\sum_{k=1}^K {\sum_{n=1}^N\sum_{s=1}^S  w(s|n) \ind{x^n_s[d] \in k}}}
\label{eq:opt:prior}
\eeq
For the M-step of the algorithm we then make a gradient update for $\theta_c$ and update the prior using \eqref{eq:opt:prior}.

\david{One could also learn the prior first, independently of learning the classifier. That is, given a set of corrupted values and knowing the corruption probabilities, learn the prior. This would still be OK since it wouldn't require an individual to release more than a single noisy version of their datapoint.}

\section{Experiments\label{sec:exp}}

We implemented our approach in \secref{sec:private:log:reg} to train a logistic regression classifier to distinguish between the MNIST digits 7 and 9 based on noisy data (250 train and 900 test examples from each class).  We chose to train on a small dataset since this constitutes the most challenging scenario and helps highlight potential differences between rival approaches. 
The MNIST images $x$ have pixesl with 256 states and we used a discrete distribution to model $x$. 

For our experiments we assume a corruption model $p(\s{c}|c)$ that flips the label $0\rightarrow 1$ and $1\rightarrow 0$ with probability $p_f$ with probability $p_f$. We also assume here for simplicity assume a factorised input corruption model $p(\s{x}|x)=\prod_{d=1}^D p(\s{x}[d]|x[d])$ 
in which with probability $1-p_f$ and uniformly from the other states of that pixel with probability $p_f$. 



\begin{figure}[t]
	\centering
	\subfloat[]{\includegraphics[height=0.1\linewidth]{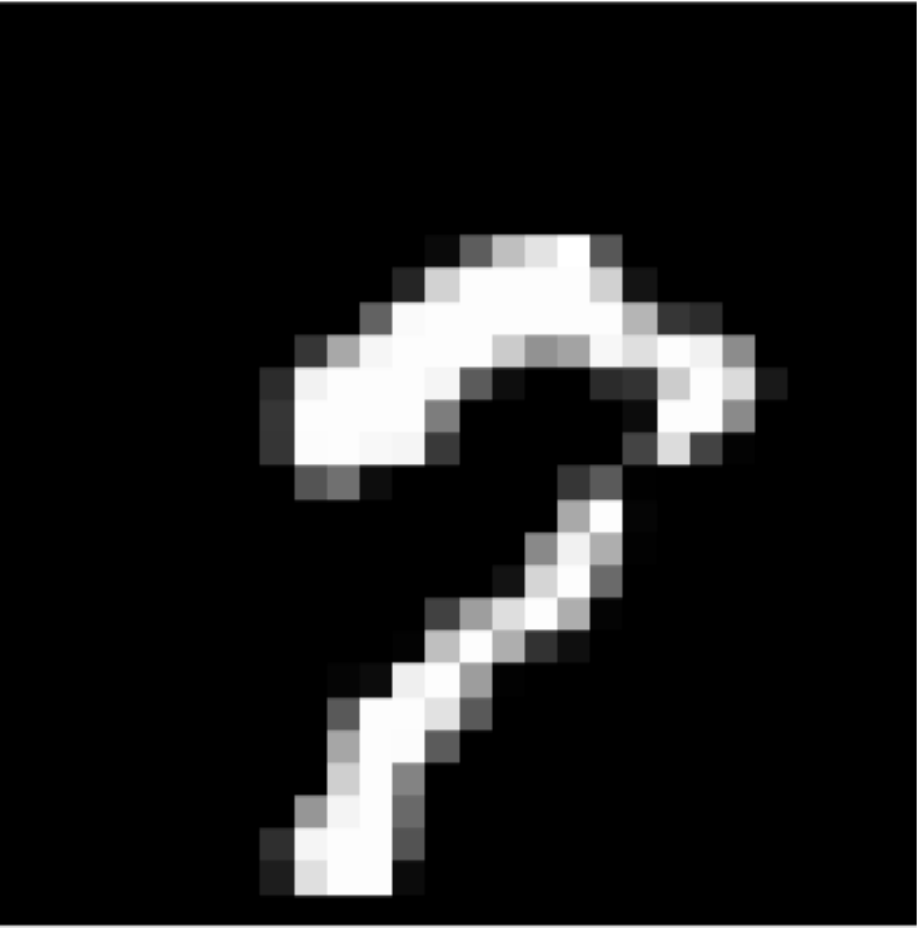}}\hspace{2mm}
	\subfloat[]{\includegraphics[height=0.1\linewidth]{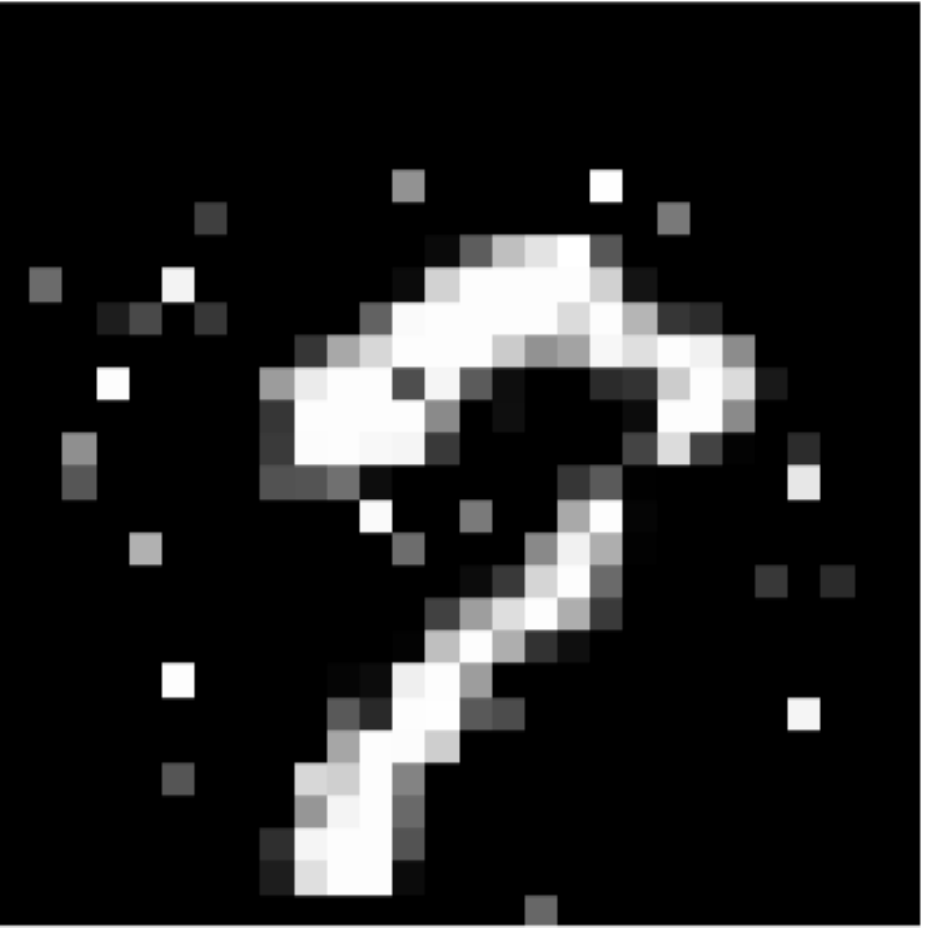}}\hspace{2mm}
	\subfloat[]{\includegraphics[height=0.1\linewidth]{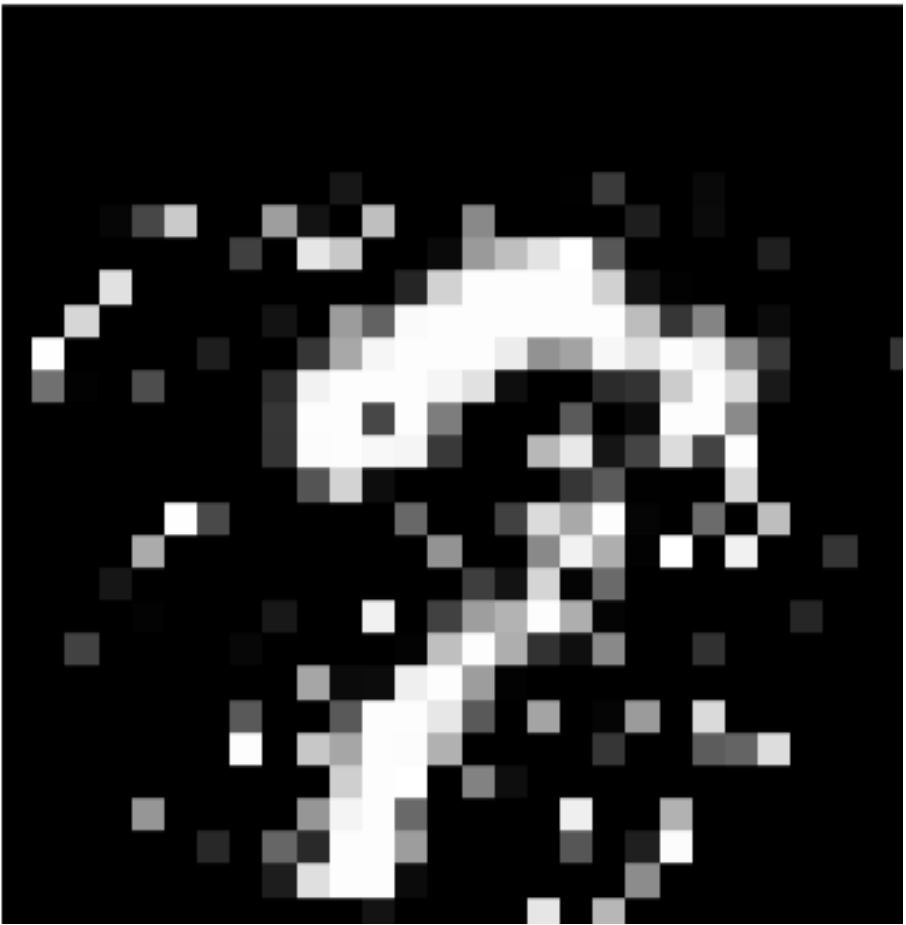}}\hspace{2mm}
	\subfloat[]{\includegraphics[height=0.1\linewidth]{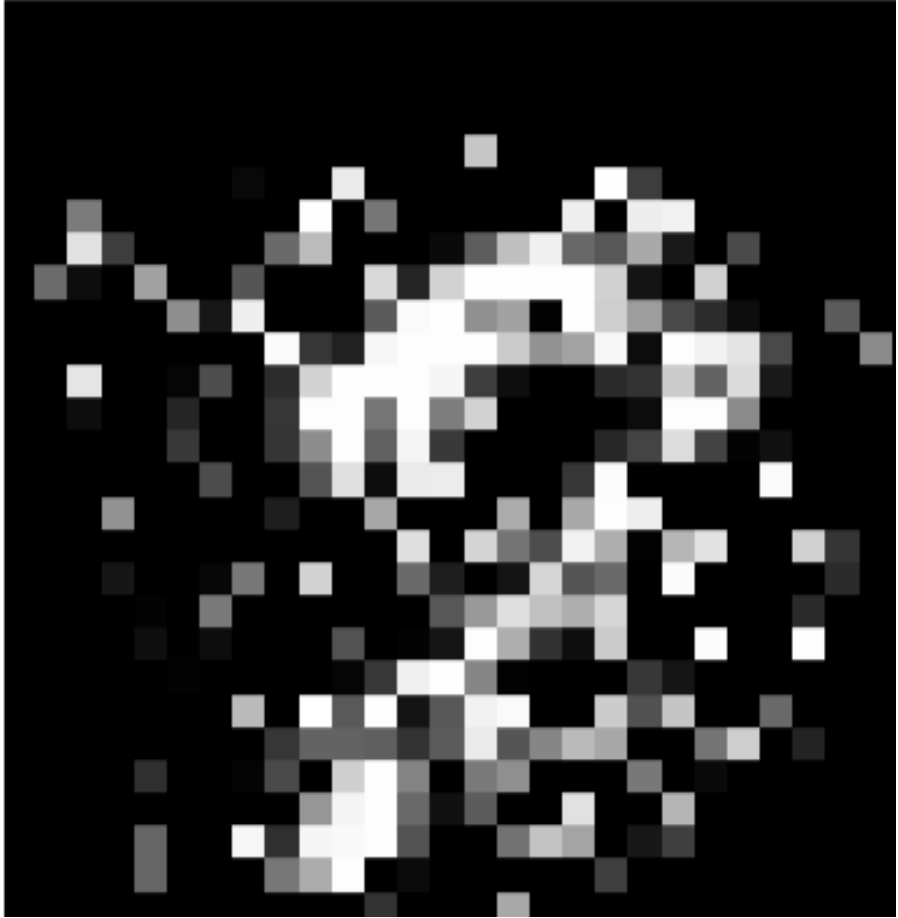}}\hspace{2mm}
	\subfloat[]{\includegraphics[height=0.1\linewidth]{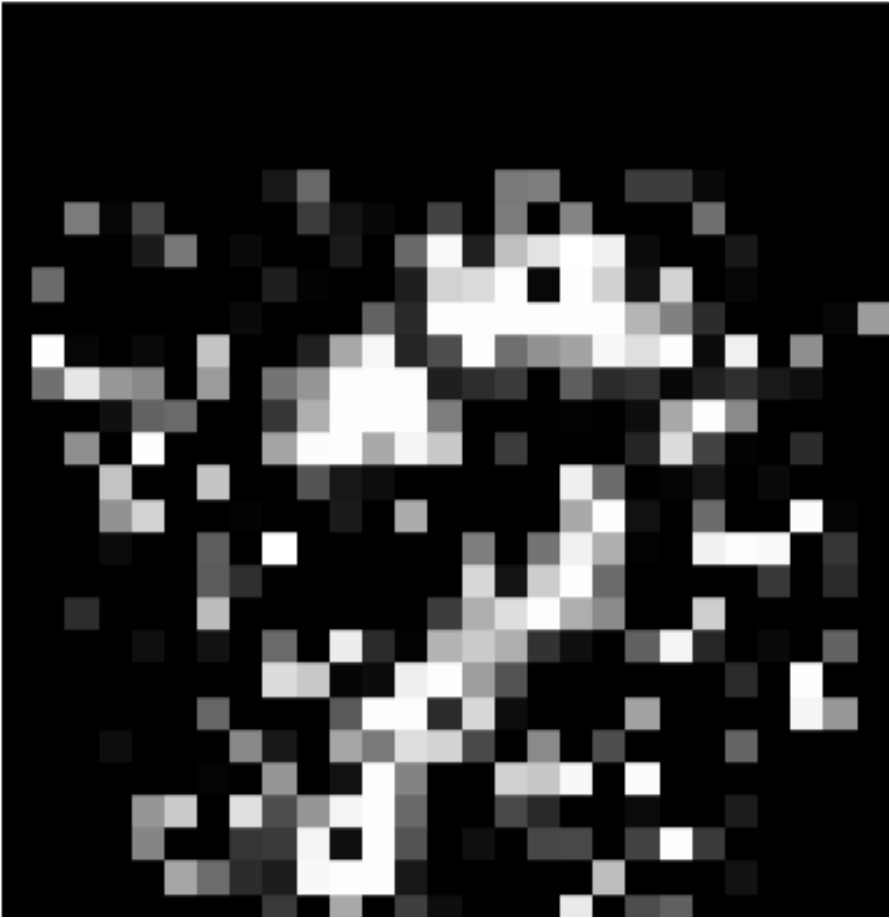}}\\
	\subfloat[]{\includegraphics[height=0.1\linewidth]{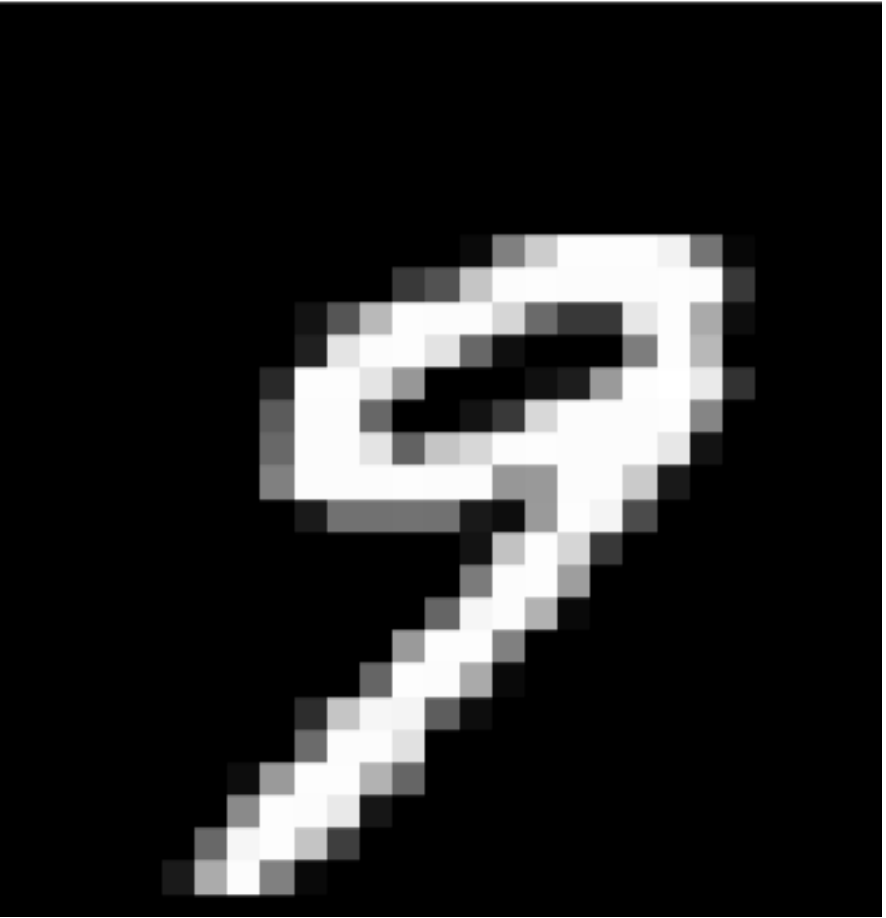}}\hspace{2mm}
	\subfloat[]{\includegraphics[height=0.1\linewidth]{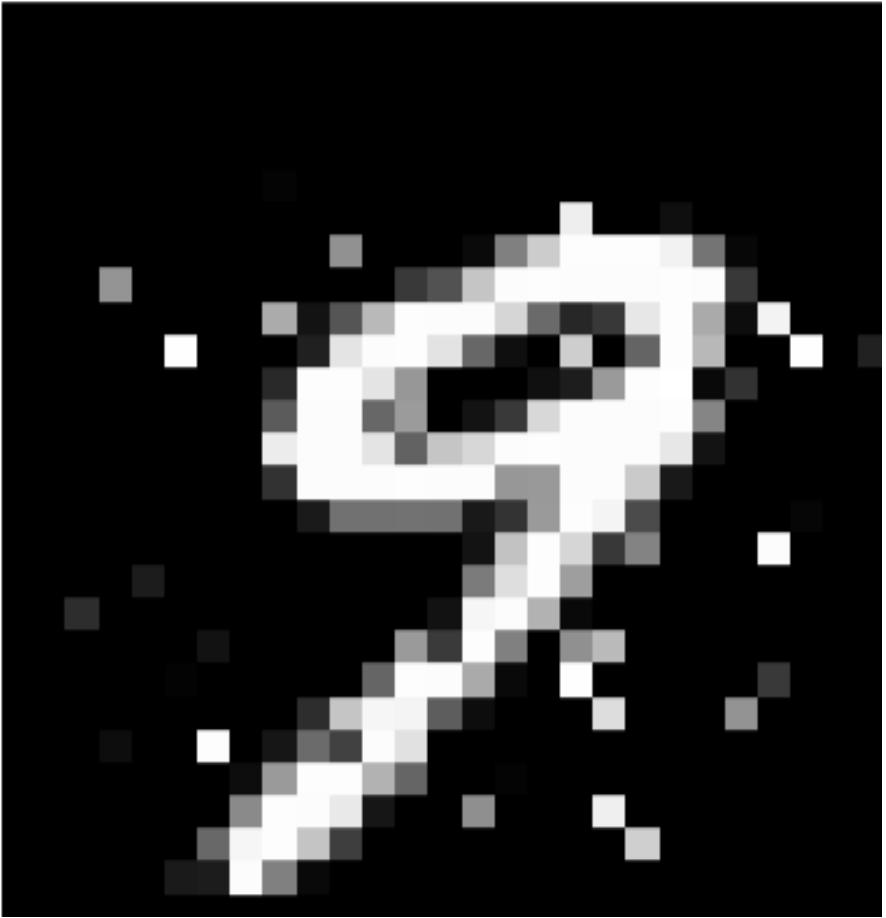}}\hspace{2mm}
	\subfloat[]{\includegraphics[height=0.1\linewidth]{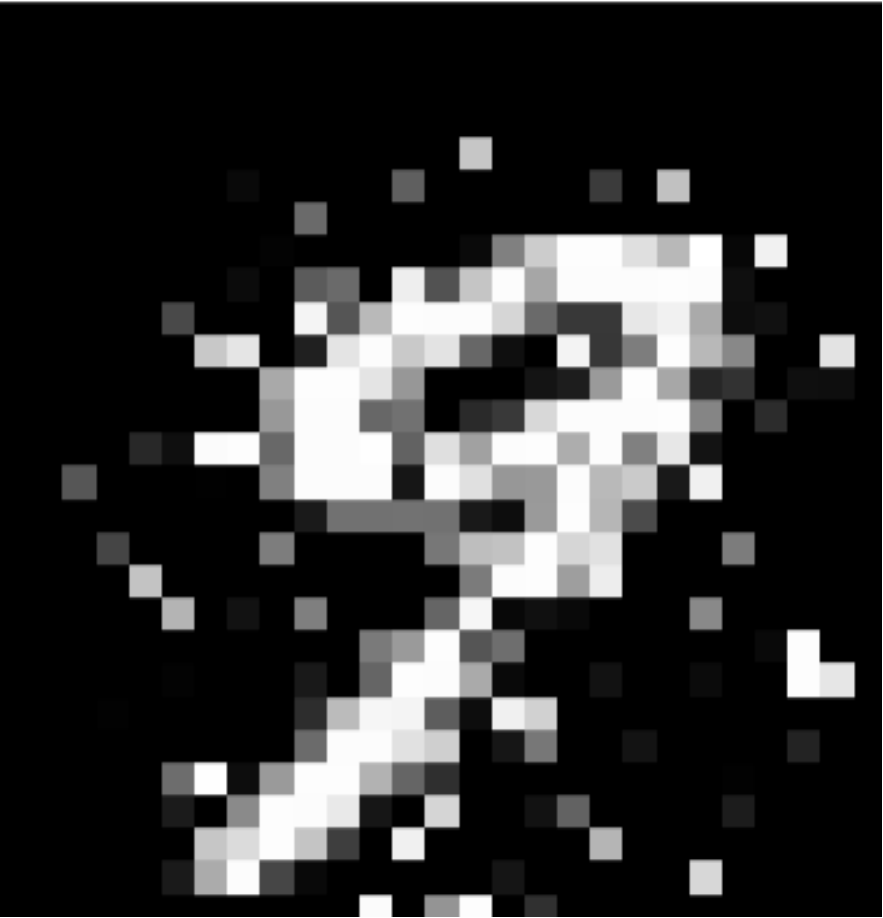}}\hspace{2mm}
	\subfloat[]{\includegraphics[height=0.1\linewidth]{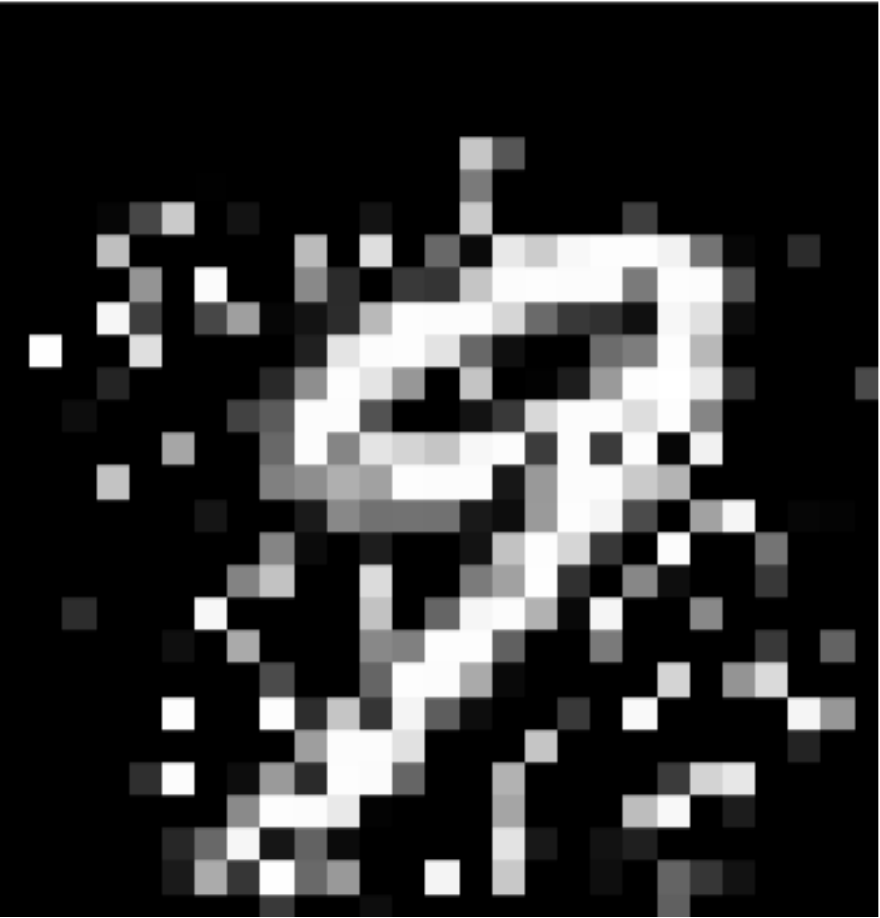}}\hspace{2mm}
	\subfloat[]{\includegraphics[height=0.1\linewidth]{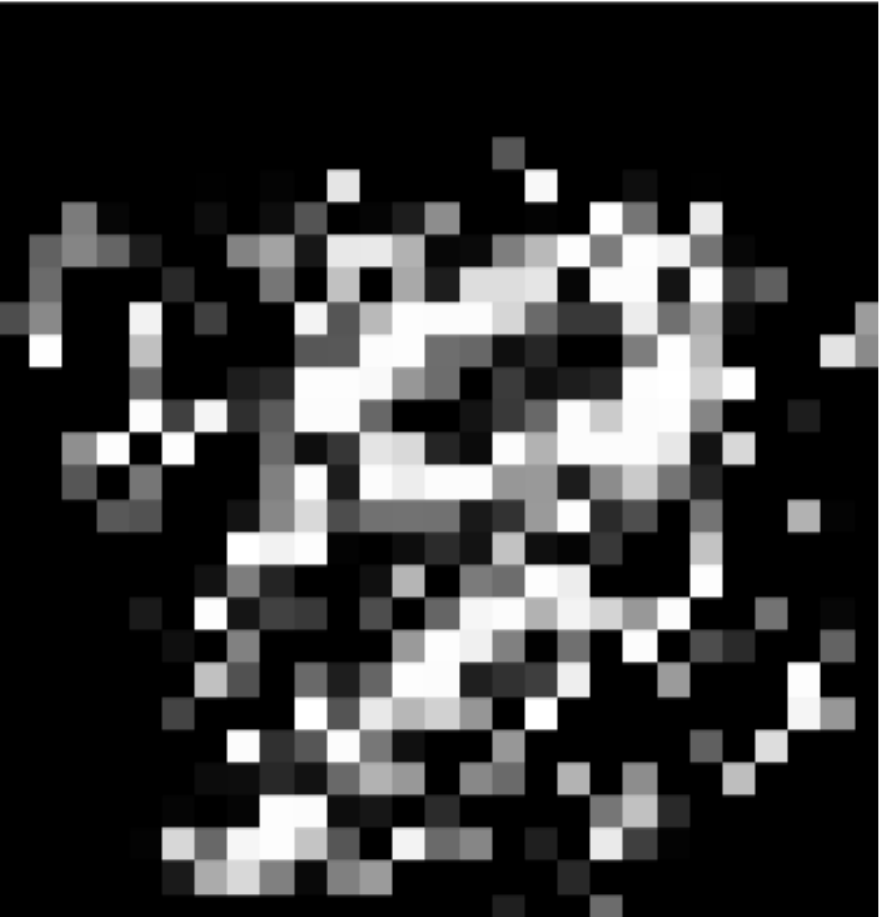}}
	\caption{(a) An example MNIST ``7'' alongside its noisy examples (b) $p_f=0.1$, (c) $p_f=0.2$, (d) $p_f=0.3$, (e) $p_f=0.4$) which is sent to Mugshot.com noise. Each pixel remains in state $1-p_f$ and is otherwise sampled uniformly from the available states of that pixel. The bottom row shows an example of a clean ``9'' (f) and corruptions.\label{fig:xtilde}}
\end{figure}

In this case, computing the Importance Sampling distribution is straightforward since the posterior is factorised over the image pixels. We considered three settings for the prior (required to compute the Importance Sampling distribution) : (i) flat prior, (ii) learned prior using EM, (iii) true factorised prior based on computing the marginal distribution of each pixel on the training dataset. In the `true prior' case we assumed that we know the true marginal distribution of each pixel $p(x[d]|d)$ -- in general, this information would be private, but it is interesting to consider how much improvement is obtained by knowing this quantity.

We compare the following approaches:
\begin{description}
	\item[Log Reg Clean Data] We trained logistic regression on clean data. This sets an upper limit on the expected performance. 
	\item[Log Reg on Noisy Data] We trained a standard logistic regression model but using the corrupted data. This forms a simple baseline comparison.
	\item[Spread Log Reg with Learned Prior] We used our Spread Likelihood approach to learn the prior.
	\item[Spread Log with `True Prior'] In general our assumption is that the true prior will not be known (since this requires users to release their private data to the prior learner). However, this forms an interesting comparison and expected upper bound on the performance of the spread approach.
	\item[Spread Log with Flat Prior] In this case we used an informative, flat prior on all pixel states.
\end{description}

\begin{figure}[t]
	\includegraphics[width=0.65\textwidth]{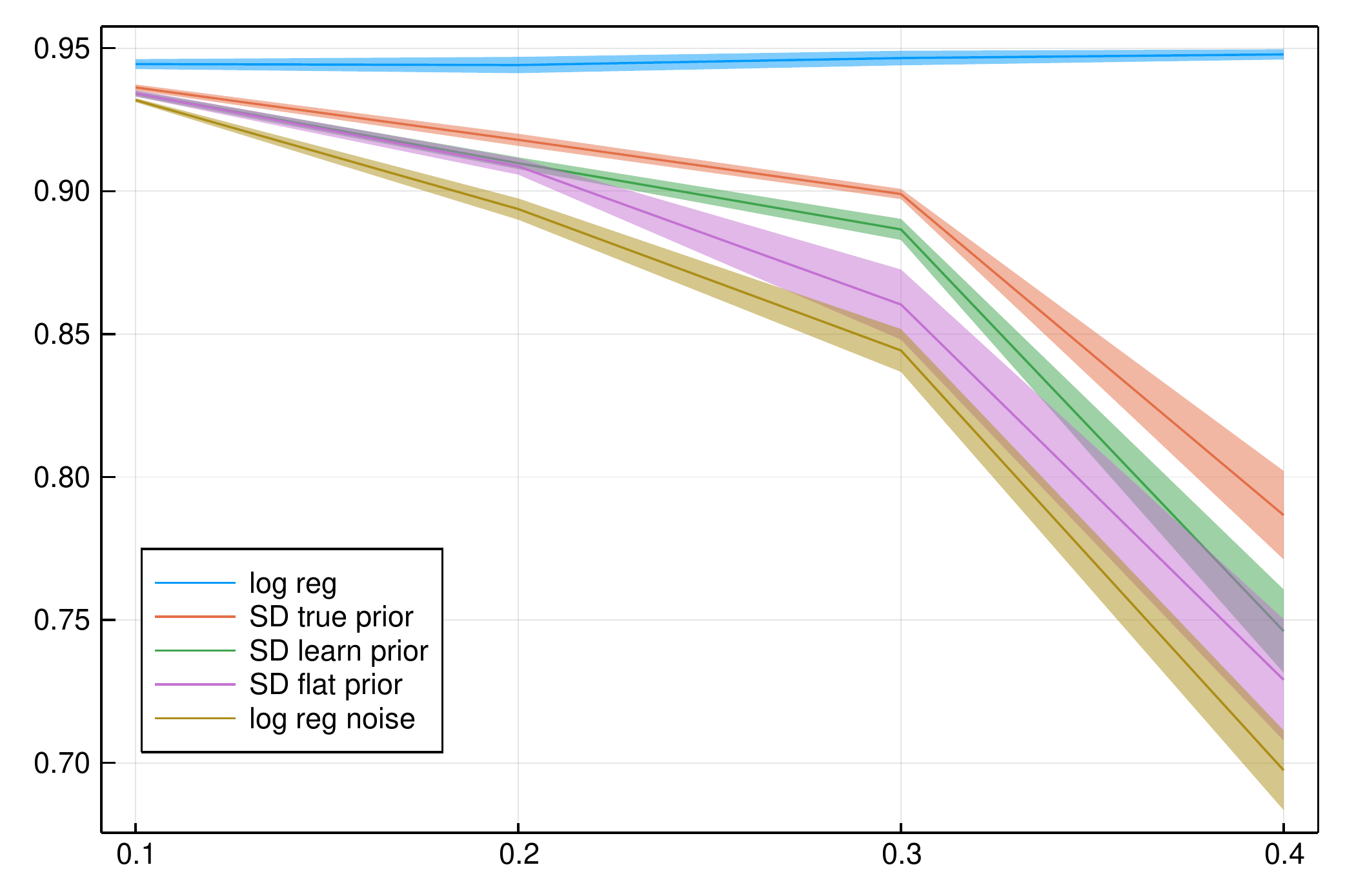}
	\caption{The test accuracy (on clean data) of the trained logistic regression models, averaged over 10 different randomly chosen training datasets of 500 datapoints. The x-axis is the corruption probability $p_f$. ``log reg'': Standard logistic regression training on clean data; ``SD true prior'': spread divergence training approach with true prior; ``SD learn prior'': spread approach with learned prior; ``SD flat prior'': spread approach with flat prior; ``log reg noise'': standard logistic regression training but trained on noisy data. \label{fig:results}}
\end{figure}

We ran 10 experiments for each level of flip noise $p_f$ from $0.1,0.2,0.3,0.4$ and then tested the prediction accuracy of the learned logistic classifiers on clean hold out data, see \figref{fig:results}. 

For all but small noise levels, the results show the superiority of the spread learning approach over simply training on noisy data, consistent with our theory that training the standard model on noisy data does not in general form a consistent estimator. The best performing spread approach is that which uses the true prior -- however, in general this true prior will not be available. For this experiment, there appears to be little difference between using a flat prior and a learned prior. 

The performance of standard logistic regression training but using corrupted data is surprisingly effective, at least at low noise levels. However the performance degrades quickly for higher noise levels. A partial explanation for why logistic regression may give good results simply trained on noisy data is given in \secref{sec:train:on:noise:only}.

%
%
%
%

\subsection{Gaussian Input Prior $p(x)$}

We also demonstrate here training logistic regression treating the pixels as continuous. If we an independent have (per pixel) a Gaussian prior 
\beq
p(x_i) = \ndist{x_i}{\mu_i}{\bar{\sigma}_i^2}
\eeq
and independent Gaussian spread noise
\beq
p(\s{x}_i|x_i) = \ndist{\s{x}_i}{x_i}{\sigma^2_i}
\eeq
then using the Importance Sampling posterior is
\beq
\rho(x_i|\s{x}_i)  = \ndist{x_i}{mean=\frac{b_i}{a_i}}{var=\frac{1}{a_i}}
\eeq
where
\beq
a_i = \frac{1}{\sigma^2_i} + \frac{1}{\bar{\sigma}_i^2}, \ocm b_i = \frac{\s{x}_i}{\sigma^2_i} + \frac{\mu_i}{\bar{\sigma}_i^2}
\eeq

We also used Gaussian spread noise to corrupt the images and train a binary classifier to distinguish between the MNIST digits 7 and 9 based on noisy data (4500 train and 900 test examples from each class). For simplicity, we assumed factorised distributions with prior $p(x_i)=\ndist{x_i}{\mu_i}{\bar{\sigma}^2_i}$, $p(\s{x}_i|x_i)=\ndist{\s{x}_i}{x_i}{\sigma^2_i}$. We chose spread flip noise $p_f=0.2$ for the class labels and uniform spread noise with variance $\sigma^2_i=0.1$; the prior $p(x)$ was set to be quite uninformative with mean $\mu_i=0$ and variance $\bar{\sigma}^2_i=10$.  This level of noise means that approximately 20\% of the class labels are incorrect in the data passed to MugTome and the associated image is significantly blurred, see \figref{fig:xtilde}. For standard logistic regression we found that for a learning rate of 0.2, 400 iterations gave the best performance, with 95.5\% train accuracy and 95.7\% test accuracy.   Using our Importance Sampling scheme with $S=2$ samples per noisy datapoint, the trained $\theta$ when tested on clean images had 94.4\% test accuracy.  This shows that despite the high level of class label and image noise, MugTome are able to learn an effective classifier, preserving the privacy of the users. The loss in test and training accuracy, despite this high noise level is around a modest 1\%.   When using higher spread noise with variance $\sigma^2=0.5$, the $\theta$ learned on the noisy data had a clean data test accuracy 93\%, which is also a modest decrease in accuracy for a significant increase in privacy.

\begin{figure}[b]
	\centering
	\subfloat{\includegraphics[width=0.15\linewidth]{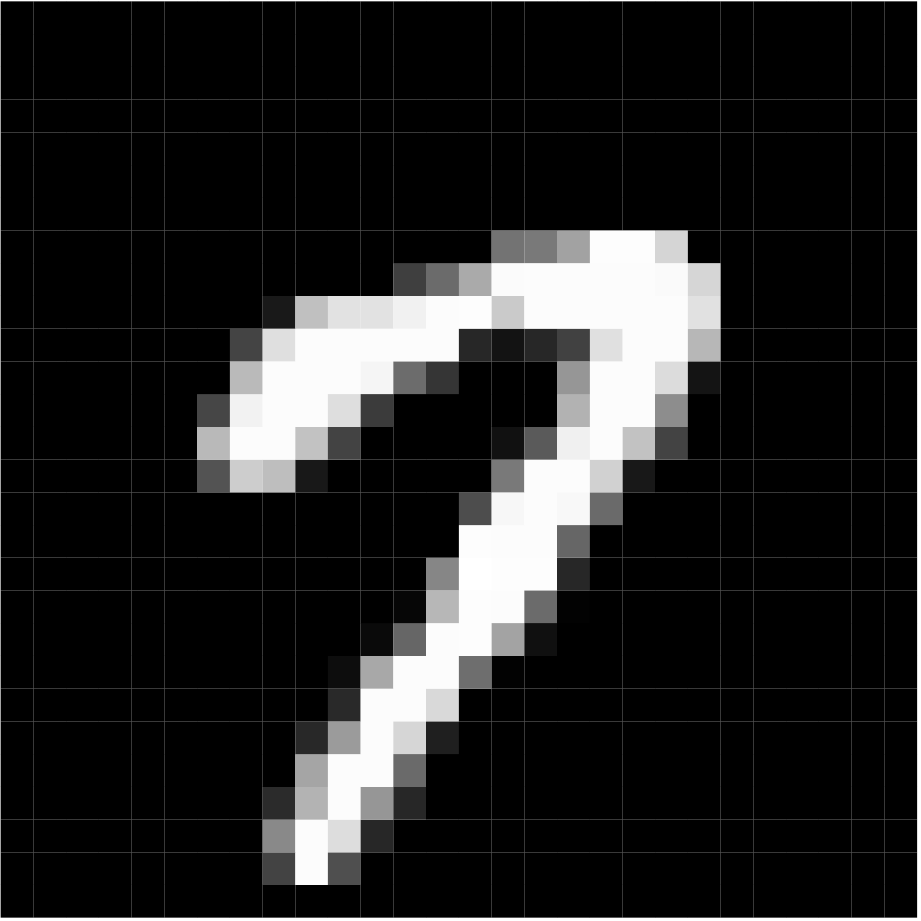}}\hspace{1mm}
	\subfloat[]{\includegraphics[width=0.15\linewidth]{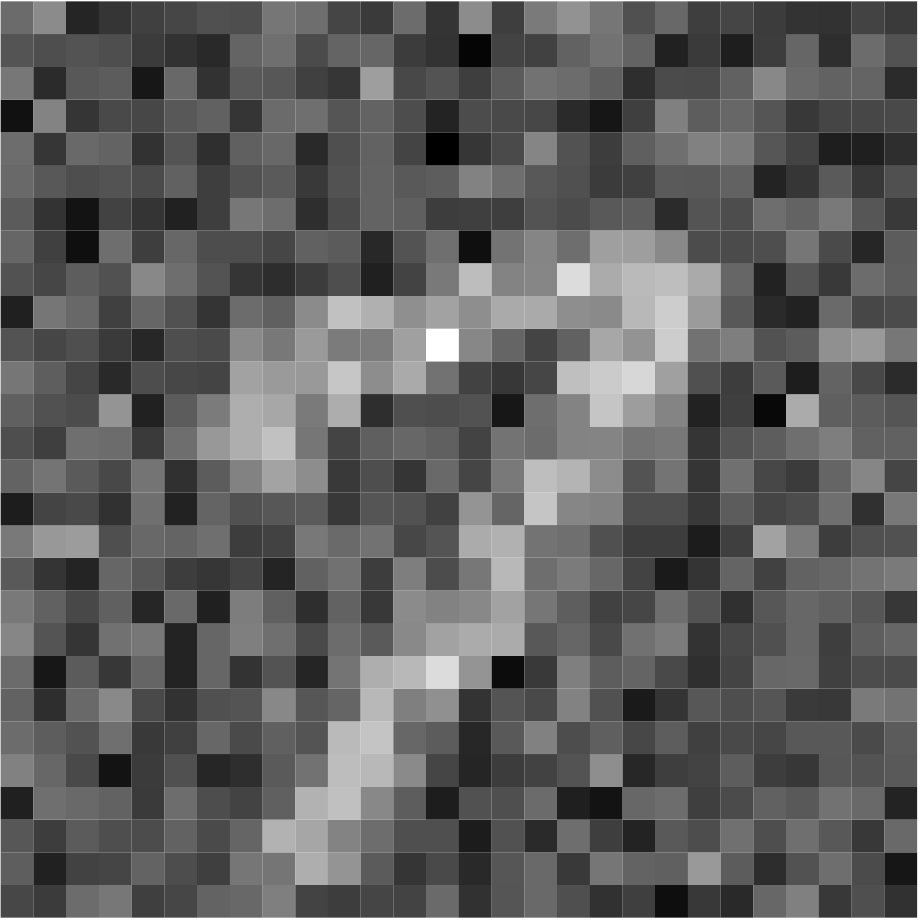}}\hspace{1mm}
		\subfloat[]{\includegraphics[width=0.15\linewidth]{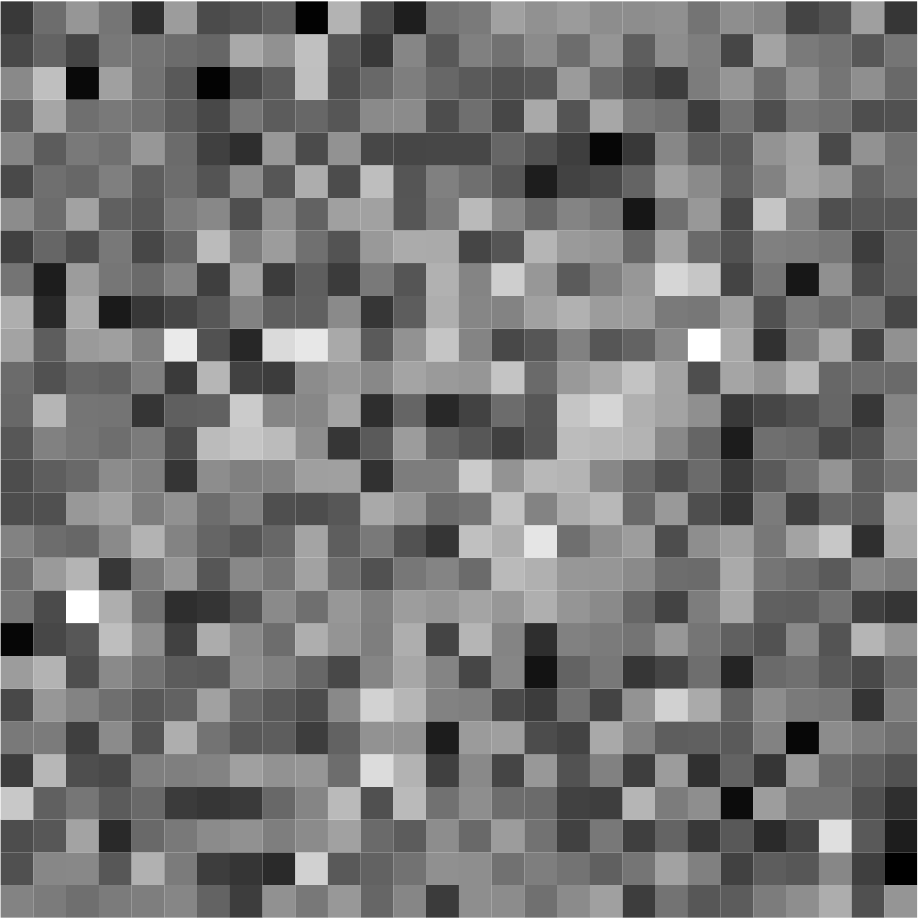}}\hspace{1mm}
	\subfloat[]{\includegraphics[width=0.15\linewidth]{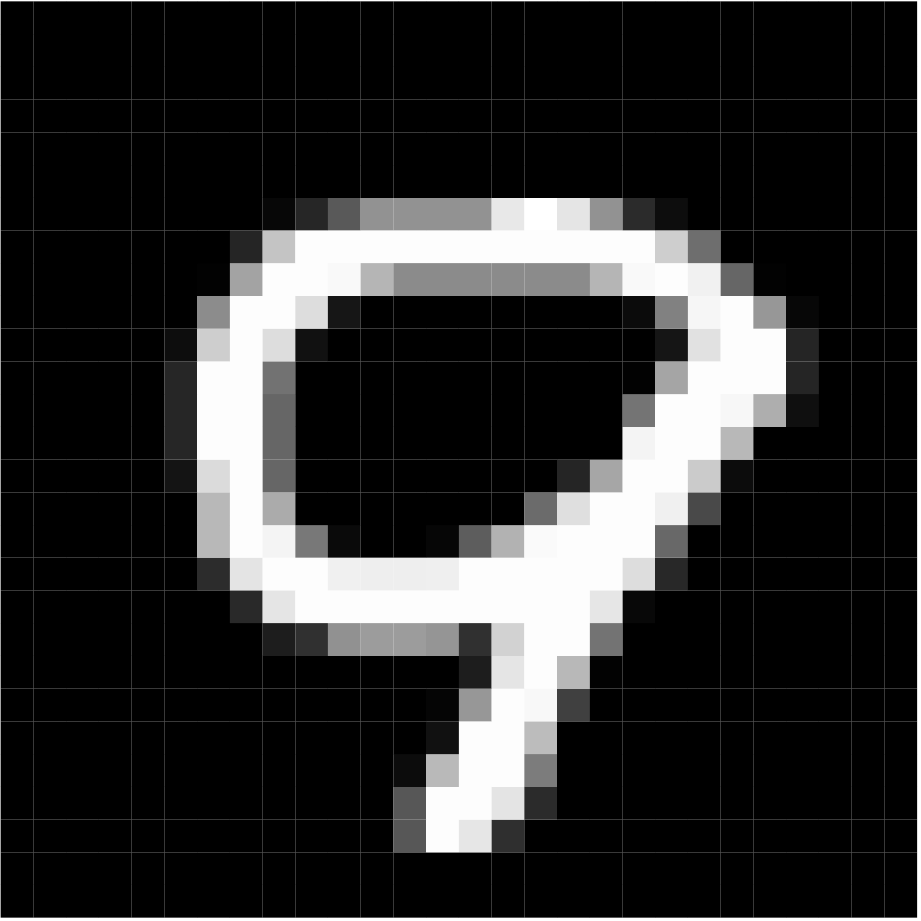}}\hspace{1mm}
	\subfloat[]{\includegraphics[width=0.15\linewidth]{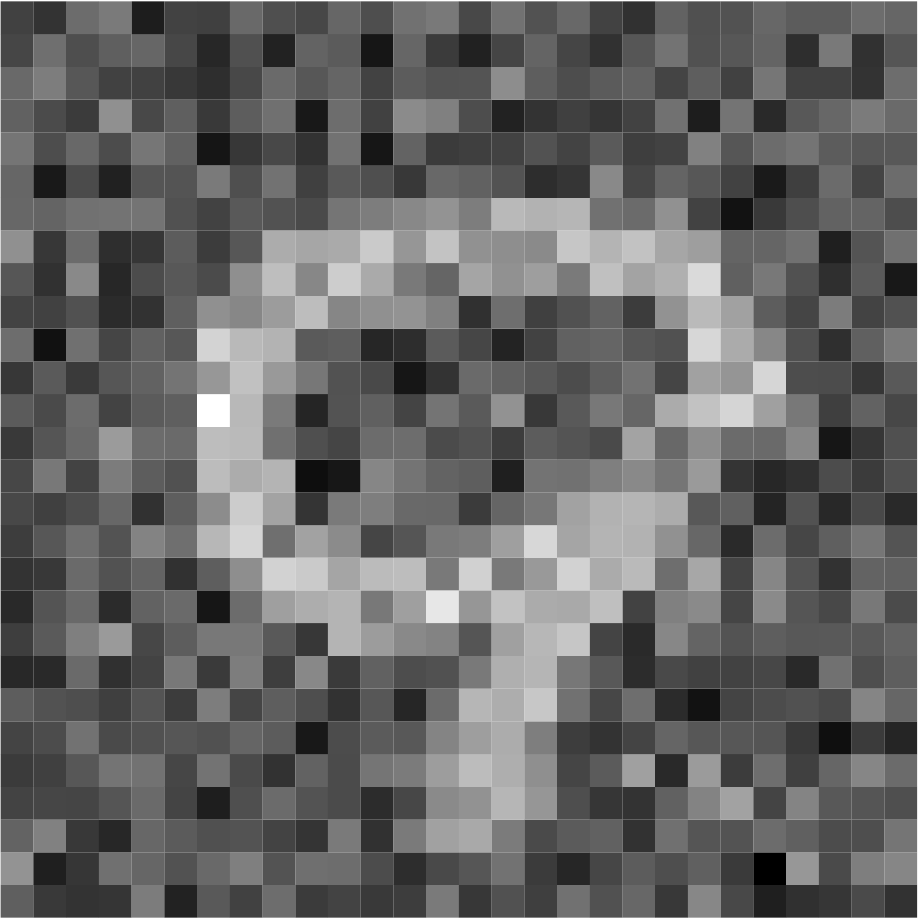}}\hspace{1mm}
		\subfloat[]{\includegraphics[width=0.15\linewidth]{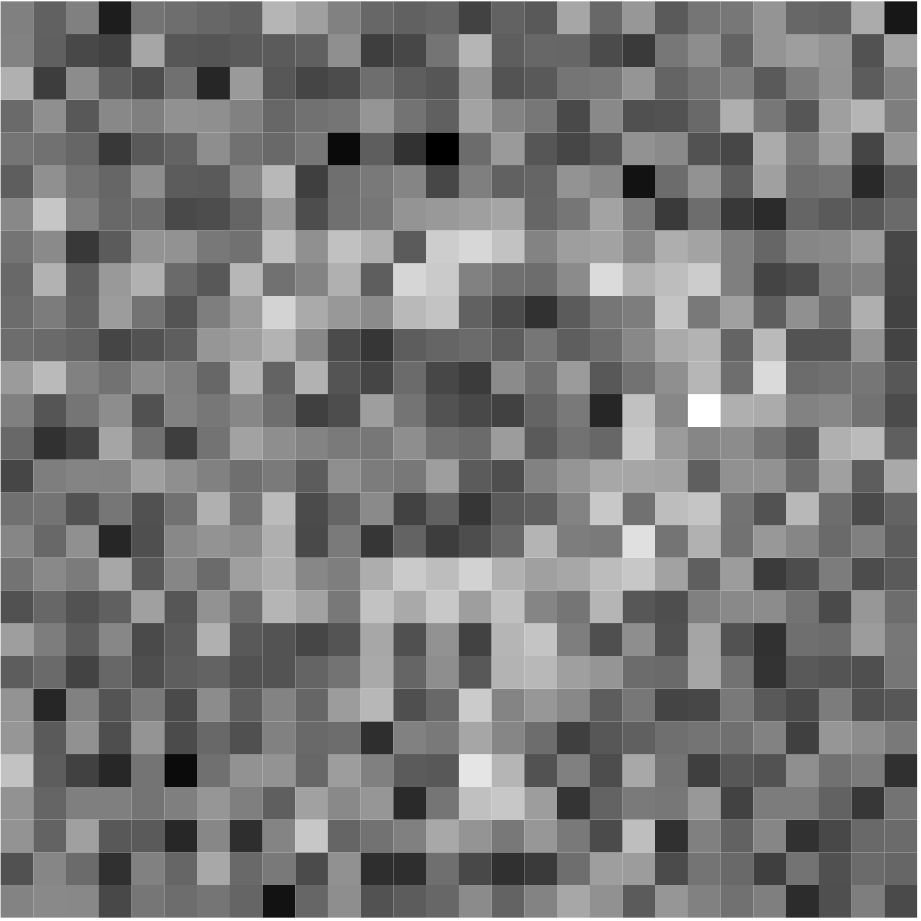}}
	\caption{(a) An example MNIST ``7'' alongside its noisy example which is sent to MugTome with Gaussian noise variance (b) $\sigma^2=0.1$ and (c) $\sigma^2=0.5$ ; (d,e,f) similarly for an MNIST ``9''. \label{fig:xtilde}}
\end{figure}

For future work it would be interesting to consider other forms of noise, for example downsampling images. However, downsampling does not form an injective mapping and as such we cannot guarantee that we can find a consistent estimator for the underlying model.

\section{Discussion\label{sec:related}}

\david{To be done. Need a comprehensive survey somewhere about existing privacy approaches.}

There are many forms of private machine learning.  Some attempt to transform a datapoint $x$ to a form $x'$ such that a protected attribute $a$ (such as gender) cannot be recovered from $x'$, yet the prediction of an output $y$ (for example using $p(y|x')$) is retained.  For example this could be achieved by using a loss function such as (see for example \cite{li2019deepobfuscator})
\beq
L(\theta,\phi,\psi) = \sum_n \sq{L_y(y_n,y(x'_n;\theta)) - L_a(a_n,a(x'_n;\phi))}
\eeq
where $n$ is the data index, $y(x';\theta)$ is a function that takes input $x'$ and outputs a prediction $y$; $a(x';\phi)$ is a function that takes input $x'$ and outputs an attribute prediction $a$ and $x'=f(x;\psi)$ gives a representation of the input; $L_y$, $L_a$ are loss functions.  In this protected attribute setting, typically some form of the clean dataset is required to learn the parameters $\theta,\phi,\psi$. 

Another common form of private machine learning is based on differential privacy \citep{Dwork:2014}, with the aim to make it difficult to discern whether a datapoint $x_n$ was used to train the predictor $y(x;\theta)$. That is, given a trained model, differential privacy attempts to restrict the ability to differentiate whether any individual's datum was used to train the model.
%
%

A closely related concept to randomised response is that of plausible deniability, namely privately corrupting a datapoint $x_n$ such that no-one (except the datapoint provider) can confidently state what the original (private) value of $x_n$ is. Recently \cite{DBLP:journals/corr/abs-1708-07975} used this to create synthetic datapoints, which were subsequently used with a standard machine learning training approach. The authors showed that generating synthetic data $\tilde{x}$ from a distribution $p(\tilde{x}|x)$ that takes dependency amongst the elements of the vector $x$ results in better machine learning predictors than sampling from a factorised distribution. 
In synthetic data generation approaches the assumption is that the statistical characteristics are similar to the real data. However, care is required since if the generating mechanism is powerful, it may generate data which is very similar to the private data. 

In general these synthetic data generating approaches do not take into consideration when learning the parameters of the machine learning model what that synthetic data generation mechanism is.  This is analogous to simply using the corrupted votes to directly estimate the fraction of voters that voted for a candidate, \eqref{eq:noisy:f}, rather than using knowledge of the data generation approach, \eqref{eq:noisy:betterf}.

\david{Need to read also \cite{NIPS2014_5392}}

\section{Summary}

We discussed a general privacy preserving mechanism based on random response in a datapoint is replaced by a corrupted versions. We showed that, provided the corruption process is a valid spread noise, then a maximum likelihood approach forms a consistent estimator. That is, even though the model is only trained on corrupted, synthetic data, it is possible to recover the true underlying data genering mechnanism on the clean data.  We applied this approach to a simple logistic regression model, showing that the approach can work well, even with high levels of noise. The approach is readily applicable to a large class of much more complex models and other divergences. 

\subsubsection*{Acknowledgements}

I would like to thank Xijie Hou for useful discussions.

\bibliography{iclr2020_conference}
\bibliographystyle{iclr2020_conference}

\appendix

\section{Privacy Preserving Logistic Regression\label{app:logreg}}
The posterior is given by
\beq
p(c,x|\s{c},\s{x}) = \frac{p(\s{c}|c)p(\s{x}|x)p(c|x)p(x)}{\sum_c\int_x p(\s{c}|c)p(\s{x}|x)p(c|x)p(x)} =
\frac{p(\s{c}|c)p(\s{x}|x)p(c|x)p(x)}{Z(\s{c},\s{x})} 
\eeq
For the learning, we need to take expectations 
\beq
\sum_c\int_x p(c,x|\s{c},\s{x}) f(x,c)s
\eeq
We use importance sampling to approximate this expectation
\begin{align}
\sum_c\int_x p(c,x|\s{c},\s{x}) f(x,c) &= \sum_c\int_x  \rho(c|\s{c})\rho(x|\s{x}) \frac{p(c,x|\s{c},\s{x})}{\rho(c|\s{c})\rho(x|\s{x})}f(x,c)\\
&= \sum_c\int_x  \rho(c|\s{c})\rho(x|\s{x}) \frac{p(\s{c}|c)p(\s{x}|x)p(c|x)p(x)}{\rho(c|\s{c})\rho(x|\s{x})Z(\s{c},\s{x})}f(x,c)
\end{align}
Choosing
\beq
\rho(c|\s{c}) = \frac{p(\s{c}|c)}{\sum_ c p(\s{c}|c)} = \frac{p(\s{c}|c)}{Z_\rho(\s{c})}, \ocm \rho(x|\s{x}) = \frac{p(\s{x}|x)p(x)}{\int_x p(\s{x}|x)p(x)}=\frac{p(\s{x}|x)p(x)}{Z_\rho(\s{x})}
\eeq
we have
\begin{align}
\sum_c\int_x p(c,x|\s{c},\s{x}) f(x,c) & = Z_\rho(\s{c})Z_\rho(\s{x})\sum_c\int_x  \rho(c|\s{c})\rho(x|\s{x}) \frac{p(c|x)}{Z(\s{c},\s{x})}f(x,c)
\end{align}
Here
\begin{align}
Z(\s{c},\s{x}) &= \sum_c\int_x \rho(c|\s{c})\rho(x|\s{x}) \frac{p(\s{c}|c)p(\s{x}|x)p(c|x)p(x)}{\rho(c|\s{c})\rho(x|\s{x})}\\
&= \sum_c\int_x \rho(c|\s{c})\rho(x|\s{x}) \frac{p(\s{c}|c)p(\s{x}|x)p(c|x)p(x)}{\rho(c|\s{c})\rho(x|\s{x})}\\
&= Z_\rho(\s{c})Z_\rho(\s{x})\sum_c\int_x \rho(c|\s{c})\rho(x|\s{x}) p(c|x)
\end{align}
Putting this together and using the same samples to estimate the numerator and denominator expectations,
\begin{align}
\sum_c\int_x p(c,x|\s{c},\s{x}) f(x,c) & =\sum_c\int_x  \rho(c|\s{c})\rho(x|\s{x}) \frac{p(c|x)}{\sum_c\int_x \rho(c|\s{c})\rho(\s{x}|x) p(c|x)}f(x,c)\\
&\approx \sum_s \frac{p(c^s|x^s)}{\sum_s p(c^s|x^s)} f(x^s,c^s)\\
&=\sum_s w(s) f(x^s,c^s)
\end{align}
for importance weight
\beq
w(s) = \frac{p(c^s|x^s)}{\sum_s p(c^s|x^s)}
\eeq

For the logistic regression case, we have
\beq
f(x,c) = \log p(c|x) = \log\logreg\br{ (2c-1)\theta\trans{}x}
\eeq
where $\logreg(x) = 1/(1+\exp(-x))$. 

The variational lower bound then becomes
\beq
\sum_{n=1}^N  \sum_{s=1}^S w(s|n)\log\logreg\br{ (2c_n^s-1)\theta\trans{}x_n^s}
\eeq
where, for a given noisy datapoint $\s{c}_n,\s{x}_n$ we generate a set of $S$ samples $c_n^1,\ldots, c_n^S$ from $\rho(c|\s{c}_n)$ and samples $x_n^1,\ldots, x_n^S$ from $\rho(x|\s{x}_n)$  
\beq
w(s|n) = \frac{\logreg\br{ (2c_n^s-1)\theta\trans{}x_n^s}}{\sum_s \logreg\br{ (2c_n^s-1)\theta\trans{}x_n^s}}
\eeq
%
%
%
%

\section{Training on Noisy data\label{sec:train:on:noise:only}}

A common approach in private machine learning is to train the standard model based on noisy alone, corresponding to maximising $L_N'(\theta)$, \eqref{eq:log:lik:noise}. As we discussed, this does not in general give a consistent estimator of the true underlying model. To show this, we consider a logistic regression in which only the class labels $c$ are corrupted with probabilities $\pzo\equiv p(\s{c}=1|c=0)$ and in the same state with $\poo\equiv p(\s{c}=1|c=1)$, leaving the inputs $x$ uncorrupted. In this case
\beq
L'_N(\theta)=\frac{1}{N}\sum_{n=1}^N \ind{\s{c}_n=1}\log \logreg\br{\theta_c\trans x_n} + \ind{\s{c}_n=0}\log \br{1-\logreg(\theta_c\trans x_n)}
\eeq
If we assume that the true labels are drawn from an underlying model
\beq
p(c=1|x)=\logreg\br{\theta_0\trans x}
\eeq
then the probability of a corrupted label is given by
\beq
p(\s{c}=1|x)=\poo\logreg\br{\theta_0\trans x}+\pzo \br{1-\logreg\br{\theta_0\trans x}}
\eeq
and by the law of large numbers, $L'_N$ tends to
\beq
L'_\infty(\theta) \equiv \ave{ p(\s{c}=1|x)\log\logreg(\theta_c\trans x) +(1-p(\s{c}=1|x))\log\br{1-\logreg(\theta_c\trans x)) }}{p(x)}
\label{eq:large:obj}
\eeq
Taking the gradient wrt $\theta_c$, we obtain
\beq
g_\theta = \ave{ \br{p(\s{c}=1|x)(1-\logreg(\theta_c\trans x))x -(1-p(\s{c}=1|x))\logreg(\theta_c\trans x)}x }{p(x)}
\eeq
A sufficient condition for the gradient to be zero is
\beq
p(\s{c}=1|x)(1-\logreg(\theta_c\trans x)) =(1-p(\s{c}=1|x))\logreg(\theta_c\trans x)
\eeq
That is
\beq
p(\s{c}=1|x)=\logreg(\theta_c\trans x)
\eeq
which is
\beq
\poo\logreg\br{\theta_0\trans x}+\pzo\br{1-\logreg\br{\theta_0\trans x}}=\logreg(\theta_c\trans x)
\label{eq:condition}
\eeq
In general, \eqref{eq:condition} does not have a solution at $\theta_c=\theta_0$.

To understand when the objective \eqref{eq:large:obj} has an optimum, we assume the data is drawn from $p(x)=\ndist{x}{\mu}{\Sigma}$, then, defining
\beq
z_1 = \theta_0\trans x, \ocm z_2 = \theta_c\trans x
\eeq
Since $x$ is Gaussian distributed,  $z$ is also Gaussian distributed with
\beq
\ave{z_1}{} =\theta_0\trans\mu, \hcm \ave{z_2}{} =\theta\trans\mu
\eeq
\beq
\ave{z_1z_2}{}-\ave{z_1}{}\ave{z_2}{} =\theta\trans\Sigma\theta_0 
\eeq
\beq
\ave{z_1^2}{}-\ave{z_1}{}^2 =\theta_0\trans\Sigma\theta_0 
\eeq
\beq
\ave{z_2^2}{}-\ave{z_2}{}^2 =\theta\trans\Sigma\theta
\eeq
We can then write the large data limit log likelihood as a two dimensional expectation
\begin{multline}
L'_\infty(\theta) = \ave{ (\poo\logreg(z_1)+\pzo(1-\logreg(z_1)))\log\logreg(z_2)}{}\\
 +\ave{1-\poo\logreg(z_1)-\pzo(1-\logreg(z_1)))\log\br{1-\logreg(z_2)) }}{}
\end{multline}
For simplicity, consider $\mu=0$, $\Sigma=s^2I$, $\theta_0\trans\theta_0=1$, $\theta\trans\theta=1$, $\theta_0\trans\theta=\cos(\alpha)$. It is straightforward to show that in this case the gradient with respect to $\alpha$ is zero when $\alpha=0$, namely when $\theta=\theta_0$.  However, in general, for non-isotropic data covariance $\Sigma$, the gradient is non-zero at $\theta=\theta_0$.

To derive the above result, we note that the covariance for $z$ in this case is simply
\beq
C\equiv  s^2\left[ {\begin{array}{cc}
   1 & \cos\alpha \\
   \cos\alpha & 1 \\
  \end{array} } \right] 
\eeq
We now use the decomposition $C=MM\trans$, with Cholesky factor
\beq
M =  s\left[ {\begin{array}{cc}
		1 & 0 \\
	\cos\alpha & \sin\alpha \\
\end{array} } \right]
\eeq
Then drawing a sample from $z$ is equivalent to $z\sim  M\epsilon$ for $\epsilon\sim \ndist{\epsilon}{0}{I}$. Defining
\beq
\gamma(x)=\poo\logreg(x)+\pzo(1-\logreg(x))
\eeq
we can then write the expected log likelihood as a function of $\phi$:
\beq
L'_\infty(\alpha) = \ave{\gamma(Z_1(\epsilon_1))\log\logreg(Z_2(\epsilon_1,\epsilon_2))+(1-\gamma(Z_1(\epsilon_1)))\log\br{1-\logreg(Z_2(\epsilon_1,\epsilon_2))}}{\ndist{\epsilon}{0}{I}} 
\eeq 
where the functions are defined as
\beq
Z_1(\epsilon_1) = s\epsilon_1, \ocm Z_2(\epsilon_1,\epsilon_2) = s\br{\epsilon_1\cos\alpha+\epsilon_2\sin\alpha}
\eeq
Differentiating $L'_\infty(\alpha) $ with respect to $\alpha$, we obtain 
\beq
s\ave{\br{\gamma(Z_1(\epsilon_1))-\logreg(Z_2(\epsilon_1,\epsilon_2))}(\epsilon_2\cos\alpha-\epsilon_1\sin\alpha)}{\ndist{\epsilon}{0}{I}} 
\eeq
When $\alpha=0$ we note that $Z_2(\epsilon_1,\epsilon_2)$ is independent of $\epsilon_2$ and that the above is therefore is zero. Hence $\tilde{L}'_\infty(\alpha)$ has zero gradient at $\alpha=0$.

It is straightforward to show that the second derivative of $L'_\infty(\alpha)$ (evaluated at $\alpha=0$) is
\beq
s\ave{-\epsilon_1\gamma(s\epsilon_1)}{\ndist{\epsilon_1}{0}{1}} -s^2\ave{\logreg(s\epsilon_1)(1-\logreg(s\epsilon_1))}{\ndist{\epsilon_1}{0}{1}} 
\label{eq:hess}
\eeq 
The second term in \eqref{eq:hess} above is clearly negative. Using integration by parts (and noting that we may assume $s>0$), one may easily show that the first term is also negative provided that $\poo>\pzo$.

Hence we arrive at the (perhaps surprising) result that for zero mean isotropic Gaussian distributed input data, training on noisy data $(\s{c},x)$ in which the class labels have been flipped with some probability, results in a consistent estimator for $\theta_0$, provided the flip noise is not too high, namely $\poo>\pzo$, or equivalently, $\pzo+\poz<1$. This result holds even in the case of asymmetric flip noise $\pzo\neq \poz$. 

More generally, even if the data $p(x)$ is not Gaussian distributed, from the Central Limit Theorem, $p(z)$ is likely to be close to Gaussian distributed for high dimensional inputs. Hence, for input data $x$ that is roughly isotropically distributed, we can expect that training using maximum likelihood for any classifier of the form $p(c=1|x)=\logreg\br{\theta\trans x}$ will likely be close to recovering the true $\theta_0$ that generated the data (in the limit of a large number of datapoints). 

The above analysis considered only noise on the class label. If, independently of the class label we add isotropic Gaussian noise to the observations, then the projection $z$ will still be isotropic Gaussian distributed for Gaussian inputs $p(x)$ and the above argument trivially extends to this case as well. Hence, one can expect training (using standard logistic regression but with corrupted inputs and flipped labels) to be partially successful at recovering the true data generating process provided that the input data is close to isotropically distributed, motivating a whitening pre-processing step of the input data.

\end{document}